\documentclass{article}
\usepackage{ijcai22}

\usepackage{times}
\usepackage{soul}
\usepackage{url}
\usepackage[hidelinks]{hyperref}
\usepackage[utf8]{inputenc}
\usepackage[small]{caption}
\usepackage{graphicx}
\usepackage{multirow}
\usepackage{bm}
\usepackage{amsmath}
\usepackage{amsthm}
\usepackage{amsfonts}
\usepackage{booktabs}
\usepackage{subcaption}
\usepackage{array}
\usepackage{pifont}
\usepackage{xcolor}
\newcommand{\cmark}{\ding{51}}%
\newcommand{\xmark}{\ding{55}}%

\title{Self-attention on Multi-Shifted Windows for Scene Segmentation}

\author{
Litao Yu$^1$
\and
Zhibin Li$^2$
\and
Jian Zhang$^1$
\and
Qiang Wu$^1$
\affiliations
University of Technology Sydney$^1$
\and
CSIRO$^2$
\emails
\{litao.yu, jian.zhang, Qiang Wu\}@uts.edu.au$^1$
\and
zhibin.li@csiro.au
}

\begin{document}

\maketitle

\begin{abstract}
Scene segmentation in images is a fundamental yet challenging problem in visual content understanding, which is to learn a model to assign every image pixel to a categorical label. One of the challenges for this learning task is to consider the spatial and semantic relationships to obtain descriptive feature representations, so learning the feature maps from multiple scales is a common practice in scene segmentation. In this paper, we explore the effective use of self-attention within multi-scale image windows to learn descriptive visual features, then propose three different strategies to aggregate these feature maps to decode the feature representation for dense prediction. Our design is based on the recently proposed Swin Transformer models, which totally discards convolution operations. With the simple yet effective multi-scale feature learning and aggregation, our models achieve very promising performance on four public scene segmentation datasets, PASCAL VOC2012, COCO-Stuff 10K, ADE20K and Cityscapes.

\end{abstract}

\section{Introduction}

Scene segmentation is a dense classification task for visual content analysis in computer vision. The goal is to parse the objects or scenes into different 2D regions associated with semantic categories. Scene segmentation in images has drawn a broad interest for many applications such as robotic sensing \cite{ICRA14:ROBOT_SENSING} and auto-navigation \cite{ICCV09:SEG43D}. 

Recently, the development of deep convolution neural networks has led to remarkable progress in semantic segmentation due to their powerful feature representation ability to describe the local visual information. A deep segmentation network usually has the encoder-decoder learning architecture. The encoder consists of stacked convolution layers and down-sampling layers, which learns high-level semantic concepts with the progressively increased receptive field. For general scene segmentation tasks, the encoder is basically a pre-trained classification network, e.g., deep residual networks \cite{CVPR16:RESNET}. The decoder, on the other hand, aims to predict the pixel-level classes by considering the contextual visual-semantic information. In some encoder-decoder segmentation architectures such as DeepLab \cite{ECCV18:DEEPLAB}, multi-scale contextual information plays a critical role in determining the pixel-level classes, but such kinds of designs also raise an issue that they cannot learn the long-range spatial dependencies, which becomes a fundamental challenge because of the limited receptive fields applied in convolution kernels.   

\begin{figure}[t]\center
	\includegraphics[width=0.48\textwidth]{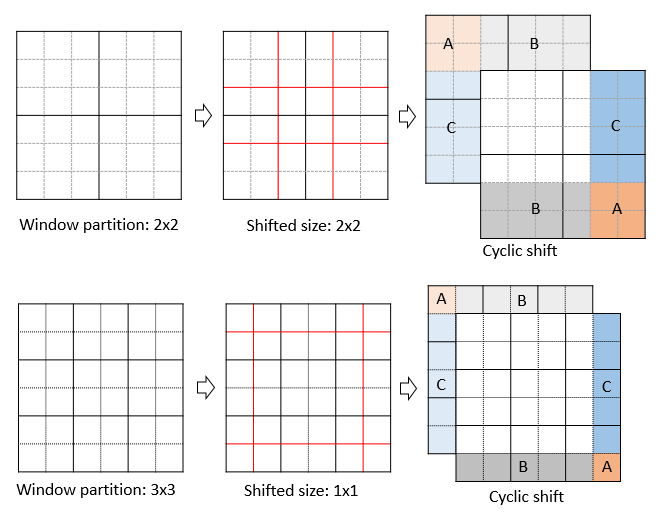}
	\caption{An illustration of multi-sifted windows. An image with $6\times6$ patches can be partitioned into $2\times2$ or $3\times3$ windows, and each partition can be separately shifted by different sizes.} \label{FIG:MSWIN6}
\end{figure}

Inspired by the success of Transformer on language modelling \cite{NIPS17:ATTN, ACL19:BERT}, the self-attention paradigm of Query-Key-Value (QKV) with multi-head attention and positional encoding has been transferred to 2D image processing, which we call as Vision Transformer (ViT). Directly partitioning an image into equal-sized patches and considering them as a sequence to learn their dependencies is a brute-force way for image classification \cite{ICLR21:VIT}. It first splits an image into equal-sized patches, then produces the low-dimensional linear embeddings. These embeddings with positional encodings are fed into the vanilla Transformer model as a token sequence. The disadvantage of this model is the lack of inductive biases of convolution neural networks, such as translation invariance and the locally restricted receptive fields, so directly training with relatively small-sized datasets obtains much inferior results. Pretrained on the very large-scale dataset, such as JFT-300M \cite{CVPR17:JFT}, then transfered to relatively small datasets, ViT achieves the state-of-the-art accuracy for classification tasks. Compared with ConvNet based scene segmentation models, the scaled dot-product attention (self-attention) in the Transformer structure has a compelling advantage because QKV learns the global dependencies of features. However, ViT does not have hierarchical feature representations. Also, the input images are supposed to be in the same size, and each image is represented by a fixed number of tokens. This is problematic for scene segmentation because the spatial-semantic consistency is difficult to be preserved at the pixel level. Also, the complexity of self-attention is quadratic to image size, leading to the comparably low efficiency. Based on the above observations, we build the scene segmentation model backend on the recently proposed Swin Transformer \cite{ICCV21:SWIN}. With the patch-merging and the self-attention within the non-overlapping windows, Swin Transformer overcomes two problems of ViT: the fixed scale tokens and the quadratic computational complexity. The pre-trained Swin Transformer itself can be used as a powerful backbone for many downstream learning tasks such as object detection and semantic segmentation. A similar work called Focal Transformer, conducts the self-attention with fine-attention locally and coarse attention globally, which enables sighificantly larger receptive fields and long-range self-attention with less memory cost and more time efficiency \cite{NIPS21:FOCAL}.

Although visual Transformers are able to learn the spatial dependencies in image processing, a simple Transformer that computes the pixel-level labels from tokens still suffers from the limitation of single-scale features, because experiences show that multi-scale information can help resolve ambiguous cases and results in more robust scene segmentation models \cite{CVPR17:PSPNET,CVPR18:DENSEASPP,ECCV18:DEEPLAB}. In Swin Transformer, the self-attention is computed within a fixed local window. Such a single-scale self-attention carries the limited local information, without the consideration of the contexts in a larger receptive field. The multi-scale feature representation is an effective way for pixel-level prediction. This motivates us to propose a Transformer with self-attention on multi-shifted windows for scale variations to decode the visual feature representations for pixel-level classifications. Specifically, we propose to use self-attentions on a series of shifted windows then aggregate them to generate visual feature maps in the decoder of the scene segmentation model. The multi-shifted window is illustrated in Figure \ref{FIG:MSWIN6}. By doing this, each neuron layer can encode the semantic information from multi-scale spatial dependencies. Therefore, the aggregated feature representations not only contain semantic information in a large scale range but also cover that range in a compact yet discriminative manner. The proposed learning framework is a pure Transformer-based scene segmentation network, which does not contain any convolution operator. We evaluate our method on four public benchmark datasets, which achieves very promising performance in terms of the mean Intersection-over-Union (mIoU) score.

We make the following contributions in this paper:

\begin{itemize}
\item We apply self-attention on multiple shifted windows then propose three feature aggregation strategies (parallel, sequential and cross-attention) in the decoder, backend on a Transformer-based pyramid feature encoder, to generate multi-scale features for scene segmentation.
\item The whole learning framework is a pure Transformer-based model, which totally discards convolution operators. Thus the computational complexity is lower than the ConvNet based segmentation models.
\item Extensive experiments show our models can learn superior feature representations as compared to fully-convolution networks (FCNs), achieving very promising performance on four public scene segmentation benchmarks. 
\end{itemize}

The rest of the paper is organized as follows. Section \ref{SEC:RELATED_WORK} introduces the related work. Section \ref{SEC:METHOD} elaborates the proposed learning framework, including a Transformer-based pyramid encoder and three decoder structures with self-attention on multi-shifted windows. Experimental results and analysis are presented in Section \ref{SEC:EXP}. Finally, Section \ref{SEC:CONCLUSION} concludes the paper. We have made our code and pre-trained models available at \textcolor{blue}{\url{https://github.com/yutao1008/MSwin}}.

\section{Related work}
\label{SEC:RELATED_WORK}

Deep learning-based image segmentation models have achieved significant progress on large-scale benchmark datasets \cite{CVPR17:ADE20K,CVPR16:CITYSCAPES} in recent years. The deep segmentation methods can be generally divided into two streams: the fully-convolutional networks (FCNs) and the encoder-decoder structures. The FCNs \cite{CVPR15:FCN} are mainly designed for general segmentation tasks, such as scene parsing and instance segmentation. Most FCNs are based on a stem-network (e.g., deep residual networks \cite{CVPR16:RESNET}) pre-trained on a large-scale dataset. These classification networks usually stack convolution and down-sampling layers to obtain visual feature maps with rich semantics. The deeper layer features with rich semantics are crucial for accurate classification, but lead to the reduced resolution and in turn spatial information loss. To address this issue, the encoder-decoder structures such as U-Net \cite{MICCAI15:UNET} have been proposed. The encoder maps the original images into low-resolution feature representations, while the decoder mainly restores the spatial information with skip-connections. Another popular method that has been widely used in semantic segmentation is the dilated (atrous) convolution \cite{ARXIV:DILATED}, which can enlarge the receptive field in the feature maps without adding more computation overhead, thus more visual details are preserved. Some methods, such as DeepLab v3+ \cite{ECCV18:DEEPLAB}, just combine the encoder-decoder structure and dilated convolution, to effectively boost the pixel-wise prediction accuracy.

With the advent of Transformer models in image processing applications, various visual Transformer models have been proposed \cite{ICLR21:VIT,ICCV21:SWIN,ICML21:DEIT}. As a central piece of Transformer, the self-attention has the complexity and structural prior challenges. The computational complexity is mainly determined by the length of tokens, so in the long-sequence modelling, the global self-attention becomes a bottleneck for model optimization and inference. In Swin Transformer \cite{ICCV21:SWIN}, the self-attention is conducted within the equal-sized window partition locally to reduce the computational complexity. Another issue of the application in 2D images is the structural prior. Unlike the invariant word embedding, the high uncertainty of image patches lead to the inductive bias, making Transformer models less effective than the convolution counterparts in computer vision tasks \cite{NIPS21:ETVT}. To obtain a comparable accuracy with ConvNet, using ViT as a backbone for image classification requires the pre-training on very large-scale datasets \cite{ICLR21:VIT}. This issue can be alleviated by applying a so-called distillation token, making the vision transformer effectively learn from a teacher (Data-Efficient Image Transformer, DeiT \cite{ICML21:DEIT}). The distillation token is learned through back-propagation by interacting with the class and patch tokens via self-attention. However, DeiT requires a pre-trained ConvNet as a teacher model. Some other techniques such as multi-stage structures \cite{ICCV21:SWIN,ICCV21:T2TVIT,ICCV21:PVT} and hybrid models \cite{ARXIV:CVT} can also make visual Transformer applicable in general image processing tasks, without the need of training on very large-scale image datasets. Among the very recently proposed visual Transformer models, Swin Transformer \cite{ICCV21:SWIN} presents a brand new perspective in designing a convolution-free deep neural network in image processing. Instead of learning the feature dependencies in the whole image patches, Swin Transformer computes the self-attention within each window partition, and uses the relative position bias $\mathbf{B}$ to each head in computing the spatial dependency:
\begin{equation}
\text{Attention}(\mathbf{Q},\mathbf{K},\mathbf{V})=\text{SoftMax}(\mathbf{QK}^{\top}/\sqrt{d}+\mathbf{B})\mathbf{V},
\end{equation}
where $\mathbf{Q},\mathbf{K},\mathbf{V}$ are query, key and value matrices, respectively. $d$ is the dimension of query and key. In each head of self-attention, the query acts as a guide to search {\em what it needs in a dictionary} to reach the final prediction. $\mathbf{B}$ is a relative position bias array rather than a scalar. In the computational pipeline, Swin Transformer first splits an image into non-overlaped patches, then applies linear mapping to embed the raw pixel values. This enables the arbitrary input sizes for Swin Transformer because the input dimension for embedding is no longer restricted by the number of pixels within a patch. The key components are Swin Transformer blocks with modified self-attention modules, which can well model the spatial dependencies within the equal-sized windows, so the computational complexity is significantly reduced. Just like convolution neural networks, the resolution reduces at the end of each stage by token merging. Also, the modified self-attention further uses a shifted window partitioning strategy, resulting in more windows to generate rich visual features thus improving the model performance. 

In the vision task of scene segmentation in images, the Transformer models also show their advantages over ConvNets counter parts. In \cite{CVPR21:SETR}, the authors deployed a pure transformer to encode an image as a sequence of patches, combined with a simple decoder to provide a powerful segmentation model SEgmentation TRansformer (SETR). Based on a hierarchical structured Transformer encoder, which outputs multi-scale features, SegFormer combined both local and global attentions for semantic segmentation \cite{NIPS21:SEGFORMER}. Yuan et. al proposed a High-Resolution Transformer \cite{NIPS21:HRFORMER}, which replaces the convolutions with local-window self-attentions in HRNet \cite{TPAMI:HRNET}. Although it achieves outstanding performance in many vision tasks, the computational complexity is very high in dense prediction. Learning feature representation at multi-scale image feature maps have been proved to be an effective way to capture the contextual information in dense prediction \cite{CVPR17:PSPNET,ECCV18:DEEPLAB,CVPR18:DENSEASPP}. So in this paper, we also follow this paradigm to design the decoder head that uses self-attention on shifted windows with different sizes. With this regard, our method achieves a satisfactory balance between computational complexity and segmentation accuracy.

\section{Method}
\label{SEC:METHOD}

\begin{figure*}[t]
  \centering
  \begin{subfigure}{0.48\linewidth}
    \includegraphics[width=1\textwidth]{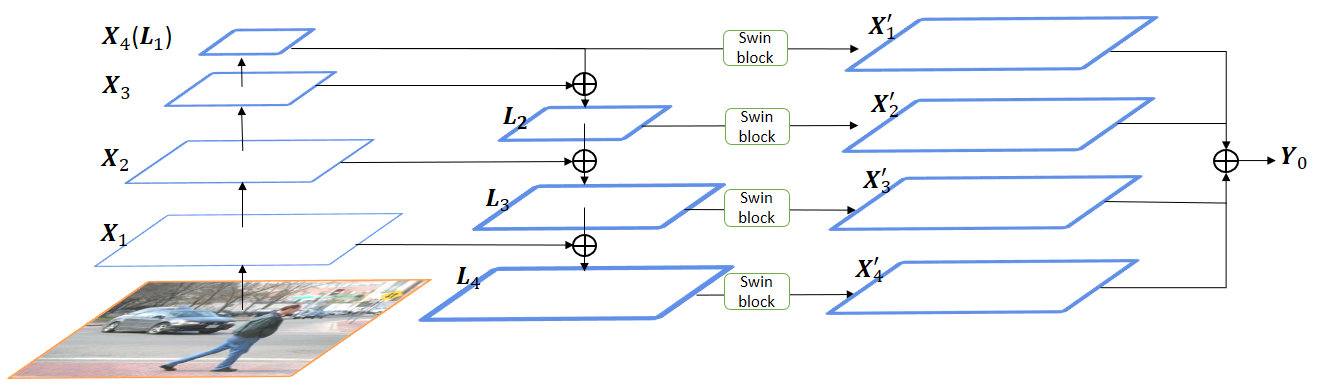}
    \caption{T-FPN.}
    \label{fig:t-fpn}
  \end{subfigure}
  \hfill
  \begin{subfigure}{0.48\linewidth}
    \includegraphics[width=1\textwidth]{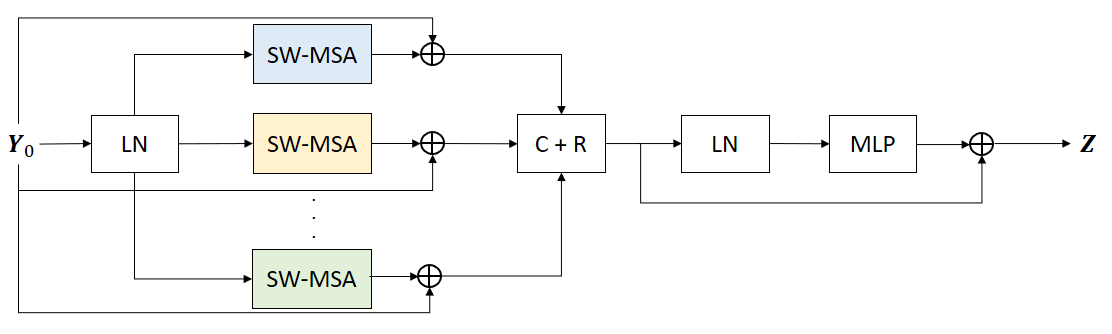}
    \caption{MSwin-P.}
    \label{fig:mswin-p}
  \end{subfigure}
  \vfill
  \begin{subfigure}{0.48\linewidth}
    \includegraphics[width=1\textwidth]{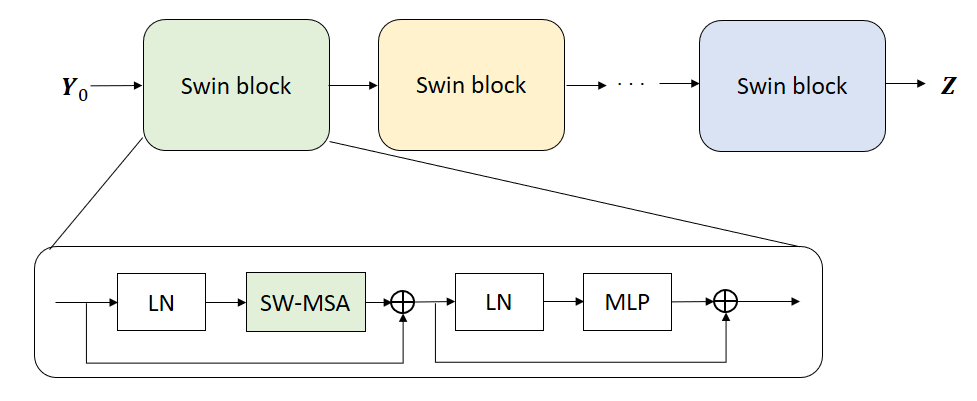}
    \caption{MSwin-S.}
    \label{fig:mswin-s}
  \end{subfigure}
  \hfill
  \begin{subfigure}{0.48\linewidth}
    \includegraphics[width=1\textwidth]{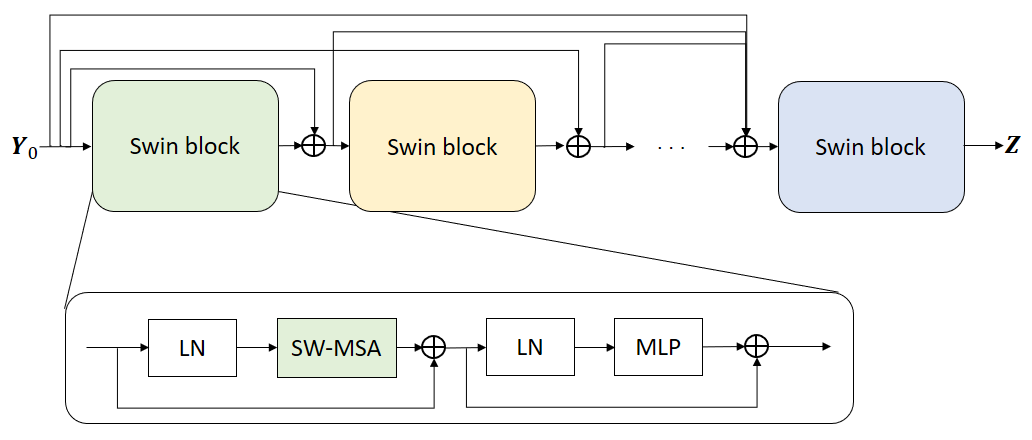}
    \caption{MSwin-C.}
    \label{fig:mswin-c}
  \end{subfigure}
  \caption{The overall learning framework of MSwin for scene segmentation. (a) The Transformer-based feature pyramid encoding structure. (b) The parallel decoding structure, where LN, C+R and MLP are layer normalization, concatenation then dimensionality reduction, and multi-layer perceptron, respectively. The SW-MSA module with different colours represent different window and shifted sizes. (c) The sequential decoding structure. (d) The cross-attention decoding structure. }
  \label{fig:frm}
\end{figure*}

In this section, we start with the design of a Transformer-based feature pyramid encoder, then introduce the decoder with self-attention on multi-shifted windows and three feature aggregation strategies. The overall learning framework is illustrated in Figure \ref{fig:frm}.

\subsection{A Transformer-based feature pyramid encoder}

The scene segmentation model is built on the recently proposed Swin Transformer \cite{ICCV21:SWIN}, which is a powerful visual Transformer in large-scale image classification and can serve as a general-purpose backbone for down-stream learning tasks in computer vision.

Swin Transformer constructs hierarchical feature maps just like the commonly used Convolution Network (ConvNet) structures such as ResNet \cite{CVPR16:RESNET} and DenseNet \cite{CVPR17:DENSENET}. However, ConvNets and Swin Transformer have fundamental differences in computing feature representations. ConvNets apply convolution operators to overlapped receptive fields to progressively learn local semantic information, while Swin Transformer learns the spatial feature dependencies within the shifted image windows. The ability of learning feature dependency in the shifted image windows endows Transformer model a compelling advantage over FCNs for scene segmentation, since it explicitly describes the spatial dependencies. Another important difference is the resolution reduction between the learning stages. ConvNets use either doubled strides or down-samplings, while Swin Transformer adopts the patch-merging scheme. The above two differences make it inappropriate to directly build an encoder with Swin Transformer just like FCNs, i.e., keeping the feature resolution in the deeper stages and increasing the dilation rate with atrous convolution to enlarge the receptive field \cite{CVPR17:PSPNET, CVPR19:DANET}. The reasons are illustrated as follows:

\begin{itemize}
\item The learned embedding of merged patches no longer provides the accurate representation for the down-streamed computations, if keeping the high-resolution features in the deeper stages;
\item If we calculate the dependency of non-adjacent patches like the way of atrous convolution \cite{CVPR17:PSPNET,ECCV18:DEEPLAB}, the self-attentions still learn the same dependency within a window.
\end{itemize}

Based on the above observations, in the design of the encoder backend on Swin Transformer, the structure of Feature Pyramid Network (FPN) with the Pyramid Pooling Module (PPM) is used as a whole learning framework (UperNet) for scene segmentation \cite{ICCV21:SWIN}. However, the PPM is applied on the low-resolution but high-dimensional feature maps in UperNet, which is computationally expensive and ineffective in restoring the spatial details. In our work, we only use the structure of Feature Pyramid Network (FPN), which can represent hierarchical features and carry rich spatial-semantic information, but remove the PPM and add a few Swin blocks before upsampling. In deep neural networks, FPN has been successfully used for object detection \cite{CVPR17:FPN} to extract multi-scale feature maps. The Transformer-based FPN is illustrated in Figure \ref{fig:t-fpn}, which composes of both bottom-up and top-down pathways. The bottom-up pathway is the four-stage computation streamline in Swin Transformer. As we go up, the spatial resolution decreases due to the patch merging. The semantic value for each stage increases with more high-level feature representation is learned. The feature outputs from the bottom layers are in high resolution but the semantic value is limited. By contrast, the features from the top layers are in low-resolution but with rich semantics. In scene segmentation, the features with high-resolution and rich semantics are both necessary for pixel-level classification, so FPN provides a top-down pathway to construct higher resolution layers and keep the rich semantic information. 

Assume an input image is $\mathbf{X}_0\in\mathbb{R}^{H\times W\times 3}$, where $H$ and $W$ are the width and height, respectively. In bottom-up pathway in Swin Transformer, the output features of the four stages are $\mathbf{X}_1\in\mathbb{R}^{\frac{H}{4}\times \frac{W}{4}\times C}, \mathbf{X}_2\in\mathbb{R}^{\frac{H}{8}\times \frac{W}{8}\times 2C}, \mathbf{X}_3\in\mathbb{R}^{\frac{H}{16}\times \frac{W}{16}\times 4C}$ and $\mathbf{X}_4\in\mathbb{R}^{\frac{H}{32}\times \frac{W}{32}\times 8C}$, respectively, where $C$ is a constant depending on the model size of the backbone. In the top-down pathway, it progressively hallucinates higher resolution features by upsampling spatially coarser but semantically richer feature maps from higher pyramid levels. These features are then fused with the feature maps computed from the bottom-up pathway via lateral connections. The lateral outputs $\mathbf{L}_1,\mathbf{L}_2, \mathbf{L}_3$ and $\mathbf{L}_4$ in the top-down pathway are computed as follows:
\begin{align}
\mathbf{L}_1 &= \mathbf{X}_4, \nonumber \\
\mathbf{L}_2 &= \text{UP}_{\times2}(\mathbf{L}_1) + \text{LP}(\mathbf{X}_3), \nonumber \\
\mathbf{L}_3 &= \text{UP}_{\times2}(\mathbf{L}_2) + \text{LP}(\mathbf{X}_2), \nonumber \\
\mathbf{L}_4 &= \text{UP}_{\times2}(\mathbf{L}_3) + \text{LP}(\mathbf{X}_1), 
\end{align} 
where $\text{UP}_{\times2}(\cdot)$ is the upsampling operator with bilinear interpolation by 2, and $\text{LP}(\cdot)$ is the linear projection with a batch normalization followed by a ReLU activation. After that, we apply four window-based self-attention blocks on the lateral outputs, then sum them up to form the final feature output $\mathbf{Y}_0$ of the encoder, as follows:
\begin{align}
\mathbf{X}'_1&= \text{UP}_{\times8}(\text{W-MSA}(\mathbf{L}_1)),  \nonumber \\
\mathbf{X}'_2&= \text{UP}_{\times4}(\text{W-MSA}(\mathbf{L}_2)),  \nonumber \\
\mathbf{X}'_3&= \text{UP}_{\times2}(\text{W-MSA}(\mathbf{L}_3)),  \nonumber \\
\mathbf{X}'_4&= \text{W-MSA}(\mathbf{L}_4),  \nonumber \\
\mathbf{Y}_0 &= \mathbf{X}'_1+\mathbf{X}'_2+\mathbf{X}'_3+\mathbf{X}'_4, 
\end{align}
where $\text{W-MSA}$ is the Window based Multi-head Self-attention module. Different from the use of FPN for object detection in \cite{CVPR17:FPN} that predicts multi-scale object sizes with pyramid feature maps, we aggregate the multi-scale lateral features in the same feature resolution. The resolution of the final feature output $\mathbf{Y}_0$ is $\frac{H}{4}\times \frac{W}{4}$.

The encoder with FPN is similar to the Unified Perceptual Parsing Network (UPerNet) proposed in \cite{ECCV18:UPERNET}. However, in our design we remove the pyramid pooling module (PPM)\cite{CVPR17:PSPNET} and apply the Swin blocks to aggregate the feature maps, thus the proposed encoder does not contain any convolution operator. Note that the PPM needs more memory and extra $3\times3$ convolutions, which is computationally expensive. In the experiment, we prove that even with the self-attention on multi-shifted windows, the whole learning framework of MSwin is more computationally efficient than UPerNet. Using FPN as the encoder of the scene segmentation model can well suit the hierarchical backbone of Swin Transformer without hurting the internal computation structure. Furthermore, the feature output carries very rich semantic information and keeps the high resolution of features. So if we directly add a classification layer as the decoder on the top of the FPN encoder output, it can form a pure Vision Transformer model for scene segmentation. Here we name it {\em Transformer based FPN} ({\bf T-FPN}), which serves as a baseline in the experiment. Note that the T-FPN only contains the self-attention on single-shifted windows.

We next introduce how to design the decoder by exploring the spatial dependencies with self-attention on multi-shifted windows. 

\subsection{The decoders with self-attention on multi-shifted windows}

The Shifted Window based Multi-head Self-attention (SW-MSA) is a simple yet effective module in Swin Transformer to enrich the feature representation. The intuition behind SW-MSA is to introduce cross-window connections of non-overlapping windows thus improving the modelling power of W-MSA. In scene segmentation tasks, it is commonly recognized that learning on multi-scale feature representations is beneficial to take different contextual information into consideration, thus giving more accurate predictions. In our model, we also follow this strategy in the design of segmentation head by applying self-attention on multi-shifted windows.  

Here we denote $m$ and $n$ be the window size and the shifted size in a SW-MSA module, where $n<m$. W-MSA is a special case of SW-MSA when $n=0$. Note that the self-attention is learned to describe the spatial dependency among non-overlapping windows, so different settings of $m$ and $n$ essentially implement multiple self-attentions in the same image. In Figure \ref{FIG:MSWIN6} we illustrate two shifted windows, where an image feature map with the resolution $6\times6$ can be partitioned to either $2\times2$ or $3\times3$ windows. Applying two shifted sizes 2 and 1 on these partitions, we can obtain four different outputs by setting $m=3,n=0$, $m=3,n=2$, $m=2,n=0$, and $m=2,n=1$ in the SW-MSA modules, respectively. This can implement a multi-scale attention within different sized windows thus can diversify the feature map representation to improve the discriminative power of self-attention for pixel-wise classification. For simplicity, we set $n=\lfloor\frac{m}{2}\rfloor$ in the self-attention on multi-shifted windows of the decoder. Assume we have $L$ SW-MSA modules, and the feature output from the T-FPN encoder is $\mathbf{Y}_0$, the three structures for the decoder are described as follows.

\subsubsection{MSwin-P: The parallel decoding structure}

The parallel structure is a {\em wide} decoder to aggregate the features from the self-attention on multi-shifted windows. It first normalizes $\mathbf{Y}_0$ then feeds its output into $L$ SW-MSA modules in parallel, then concatenates the feature outputs and applies the dimensionality reduction, followed by normalization and MLP. Here we also use the residual connections to enhance the feature representation. The structure is illustrated in Figure \ref{fig:mswin-p}, and the details of the computation streamline are described as follows:  
\begin{align}
\mathbf{Y}_{1,l}&= \text{SW-MSA}_l(\text{LN}(\mathbf{Y}_0))+\mathbf{Y}_0, \:
l=1,\ldots, L, \nonumber \\
\mathbf{Y}_2&= \text{Linear}([\mathbf{Y}_{1,1}, \ldots, \mathbf{Y}_{1,L}]), \nonumber \\
\mathbf{Z}&= \text{MLP}(\text{LN}(\mathbf{Y}_2)) + \mathbf{Y}_2, 
\end{align}
where $[\cdot]$ is the channel-wise concatenation, and $\mathbf{Z}$ is the feature output.

\subsubsection{MSwin-S: The sequential decoding structure}

The sequential structure is a {\em deep} decoder that consists of $L$ Swin blocks with different window and shifted sizes, so $\mathbf{Y}_0$ is passed through these blocks sequentially, as is illustrated in Figure \ref{fig:mswin-s}. In each Swin block, it contains only one self-attention module with/without shifted windows. The computation in each block is formulated as follows:
\begin{align} 
\mathbf{Y}_{l,1}&=  \text{SW-MSA}_l(\text{LN}(\mathbf{Y}_{l-1}))+\mathbf{Y}_{l-1}), \nonumber\\
\mathbf{Y}_{l,2}&= \text{MLP}(\text{LN}(\mathbf{Y}_{1,1})) + \mathbf{Y}_{l,1},\nonumber\\
l&= 1,\ldots, L.
\end{align}  
So the final feature output is $\mathbf{Z}=\mathbf{Y}_{L,2}$.

\subsubsection{MSwin-C: The cross-attention decoding structure}

In this decoding structure, we slightly change the SW-MSA module by using the input as the {\em query} while only keeping {\em key} and {\em value} in each Swin block. Similar to MSwin-S, there are $L$ Swin blocks with different window and shifted sizes. The query of each Swin block is the aggregation of all previous feature outputs (see Figure \ref{fig:mswin-c}), i.e., the Swin blocks are densely connected:
\begin{align} 
\mathbf{Y}_{l}&= \text{SW-MSA}_l(\sum\limits_{i=0}^{l-1}\mathbf{Y}_{i}),
\: l= 1,\ldots, L.
\end{align}
 
This design is inspired by DenseNet \cite{CVPR17:DENSENET} and DenseASPP \cite{CVPR18:DENSEASPP}. The advantage of cross-attention is its improved information flow across different self-attention modules and gradients throughout the decoder, leading to implicit deep supervision.

\section{Experiments}
\label{SEC:EXP}
We carry out comprehensive experiments on four public benchmarks to demonstrate the efficacy of the proposed learning framework. Experimental results show that equipped with the Transformer based FPN encoder, the three decoder designs achieve similar yet promising performance in scene segmentation.

\subsection{Datasets}

{\bf PASCAL VOC 2012}\cite{IJCV:VOC} contains 20 foreground object classes and one background class. The original dataset has 1,464 and 1,449 images for training and validation, respectively. To augment the training dataset, we also use extra annotations provided by \cite{ICCV11:VOC}, so the total number of training images is 10,582. Here we do not use MS COCO dataset to pretrain the segmentation model.

{\bf COCO-Stuff 10K}\cite{CVPR18:COCO} is a subset of the complete COCO-Stuff dataset, which provides pixel-wise semantic labels for the whole scene, including both ``thing'' and ``stuff'' classes. It contains 9,000 and 1,000 images for training and validation (testing), respectively. Following \cite{CVPR18:CCFGMA,CVPR19:DANET,ECCV20:OCR}, we evaluate the segmentation performance on 171 categories (80 objects and 91 stuff) to each pixel.

{\bf ADE20K}\cite{CVPR17:ADE20K} is a challenging scene parsing benchmark, in which the images are from both indoor and outdoor environments and are annotated by 150 fine-grained semantic concepts. It contains 20,210, 2,000 and 3,352 images for training, validation and testing, respectively.

{\bf Cityscapes}\cite{CVPR16:CITYSCAPES} is an urban traffic dataset, in which the images are densely annotated by 19 classes. We train the scene segmentation model with the finely annotated 2,975 images for training and 500 images for validation, respectively. It also provides 20,000 coarsely labeled images to pre-train the segmentation models.

\subsection{Implementation details}

Our experimentation is based on the semantic segmentation package {\em mmsegmentation} \cite{GIT:MMSEG}. The MSwin models implemented in the experiment are backend on a small-sized Swin-S and a medium-sized Swin-B \cite{ICCV21:SWIN}, which are pretrained on ImageNet-1K and ImageNet-22K, respectively. The window sizes in Swin-S and Swin-B are $7\times7$ and $12\times12$. In all decoder heads that contain SW-MSA modules, we set the dimension of embedding to 512 and 8-heads attention without any change. For the attention on multi-shifted windows, we used three different window sizes $5\times5$, $7\times7$ and $12\times12$, and the shifted sizes were set to $2\times2$, $3\times3$ and $6\times6$, respectively. Considering the combinations of both W-MSA and SW-MSA, there are totally $L=6$ attention blocks. Following \cite{CVPR17:PSPNET,CVPR19:DANET}, we added an auxiliary loss head composed of a two-layer sub-network. The auxiliary loss and main loss were computed concurrently with weights 0.4 and 1, respectively. We used AdamW optimizer \cite{ICLR19:ADAMW} with the initial learning rate $6\times10^{-5}$ and weight decay 0.01 after each iteration. For image augmentation, we applied random cropping, random flipping and photometric distortion. On the Cityscapes dataset, we applied random cropping with the size $512\times1024$. On the COCO-Stuff 10K, we set a smaller cropping size $480\times480$. On the rest two datasets, the cropping sizes were set to $512\times512$. We used categorical cross-entropy as loss function and report the mean Intersection-over-Union (mIoU) score in the evaluation of scene segmentation performance. These are consistent with the optimization settings of other baselines for fair comparisons, although applying some recently proposed learning objectives such as Lov{\'a}sz-softmax \cite{CVPR18:LOVASZ_SOFTMAX} and Margin calibrated log-loss \cite{IJCV:MC} can further improve the mIoU scores. We used the mixed-precision and gradient checkpoint in the model training, which allows us to set a larger mini-batch size and can effectively save the GPU memory usage without hurting the normalization layers. Our experiments were conducted on a server equipped with two NVIDIA Tesla V100 GPU cards. 

\subsection{Result on PASCAL VOC 2012 dataset}

To demonstrate the effectiveness of attention on multi-shifted windows, we conducted an ablation study of different window-size combinations on top of the T-FPN model backend on Swin-S. We separately set the window sizes $5\times5$, $7\times7$ and $12\times12$, as well as their combinations. We tested the proposed MSwin-P and MSwin-S, and the single-scaled mIoU predictions are illustrated in Table \ref{tab:ablation}. We can observe that it is unclear which single window size obtains the best results, and applying the self-attention with two window sizes do not necessarily improve the model performance. However, when we use three different window sizes by setting $L=6$, the two MSwin models perform slightly better and have relatively stable results. It is expected that using more self-attention in MSwin model with more window sizes can further boost the segmentation performance, but the computational complexity also increases accordingly.

\begin{table}[t]
  \centering
  \begin{tabular}{|ccc|ccc|}
    \hline
    \multicolumn{3}{|c|}{MSwin-P} & \multicolumn{3}{c|}{MSwin-S} \\
	\hline    
    $L$ & Size(s) & mIoU & $L$ & Size(s) & mIoU\\
    \hline
    $L=2$ & 5 & 81.29 & $L=2$ & 5 & 81.71 \\
    $L=2$ & 7 & 81.08  & $L=2$ & 7 & 81.82 \\
    $L=2$ & 12 & 81.55  & $L=2$ & 12 & 81.56 \\
    $L=4$ & 5,7 & 80.97  & $L=4$ & 5,7 & 81.79 \\
    $L=4$ & 7,12 & 81.42  & $L=4$ & 7,12 & 81.67\\
    $L=6$ & 5,7,12 & 81.58 & $L=6$ & 5,7,12 & 81.97 \\
    \hline
  \end{tabular}
  \caption{Ablation study with different window sizes on PASCAL VOC2012 validation set.}
  \label{tab:ablation}
\end{table}

\begin{figure*}[t]\centering
\begin{subfigure}{0.2\textwidth}
	\includegraphics[width=1\textwidth]{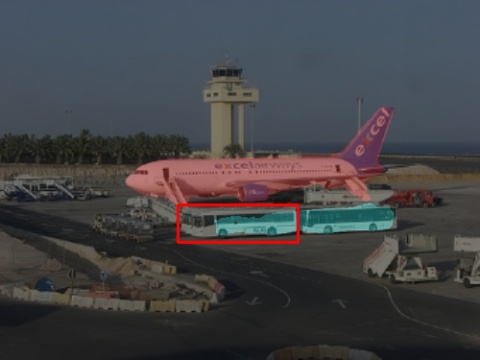}
\end{subfigure}
\begin{subfigure}{0.2\textwidth}
	\includegraphics[width=1\textwidth]{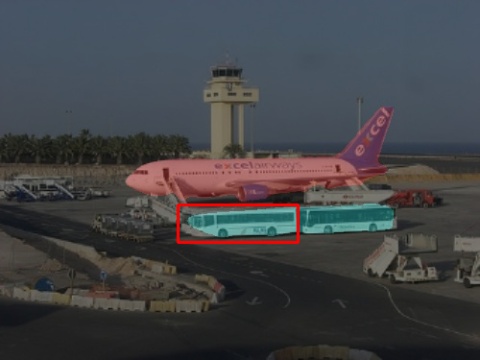}
\end{subfigure}
\begin{subfigure}{0.2\textwidth}
	\includegraphics[width=1\textwidth]{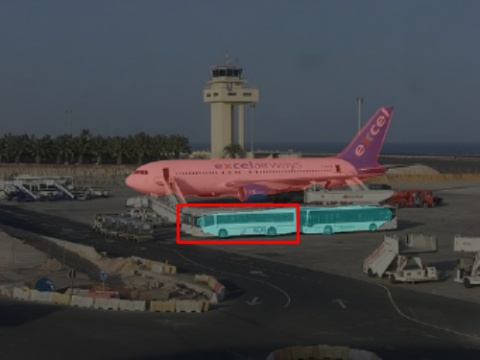}
\end{subfigure}
\begin{subfigure}{0.2\textwidth}\centering
	\includegraphics[width=1\textwidth]{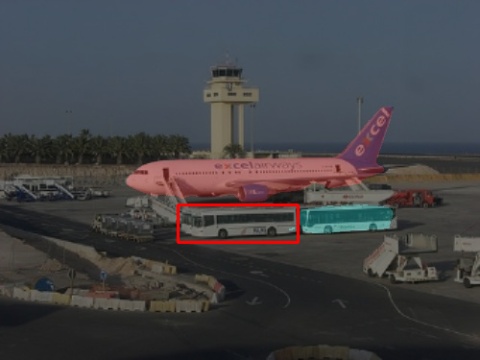}
\end{subfigure}	
\par
\begin{subfigure}{0.2\textwidth}
	\includegraphics[width=1\textwidth]{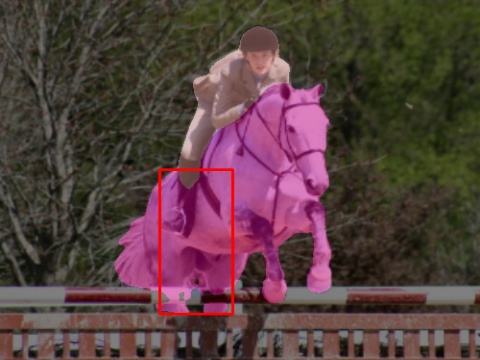}
\end{subfigure}	
\begin{subfigure}{0.2\textwidth}
	\includegraphics[width=1\textwidth]{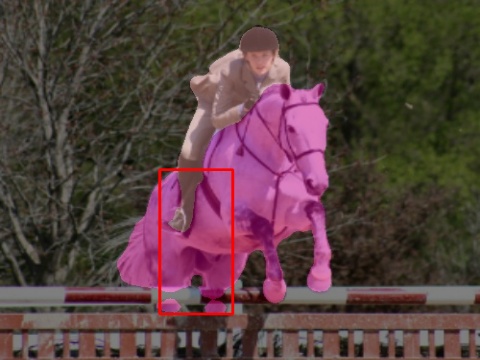}
\end{subfigure}
\begin{subfigure}{0.2\textwidth}
	\includegraphics[width=1\textwidth]{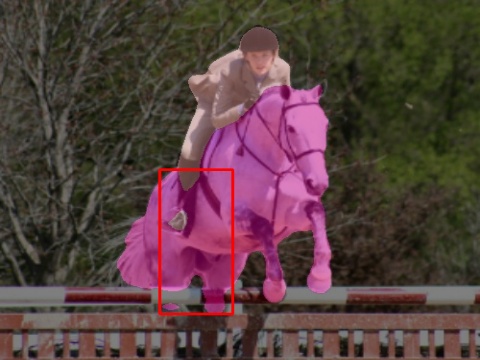}
\end{subfigure}
\begin{subfigure}{0.2\textwidth}\centering	
	\includegraphics[width=1\textwidth]{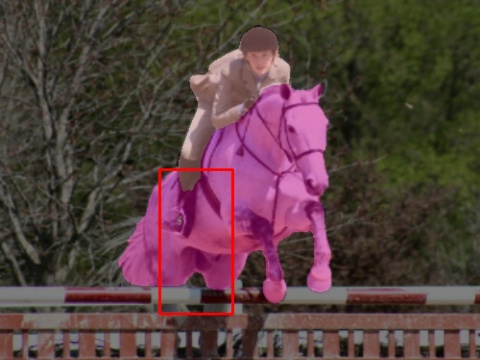}
\end{subfigure}
\par
\begin{subfigure}{0.2\textwidth}	
	\includegraphics[width=1\textwidth]{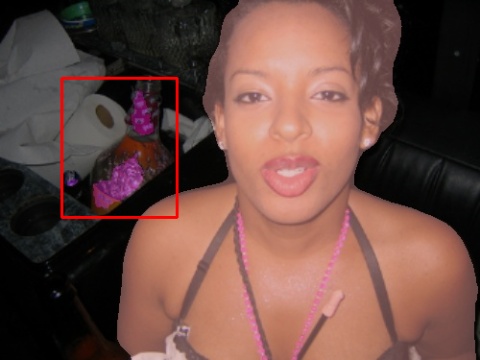}    	
\end{subfigure}
\begin{subfigure}{0.2\textwidth}	
	\includegraphics[width=1\textwidth]{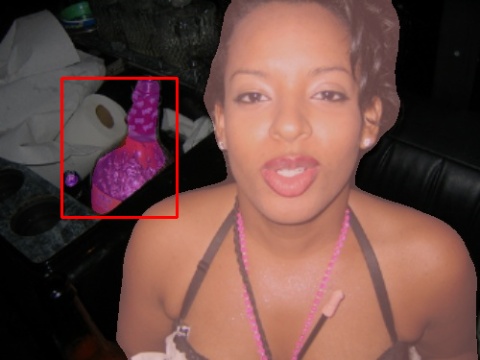}    	
\end{subfigure}
\begin{subfigure}{0.2\textwidth}	
	\includegraphics[width=1\textwidth]{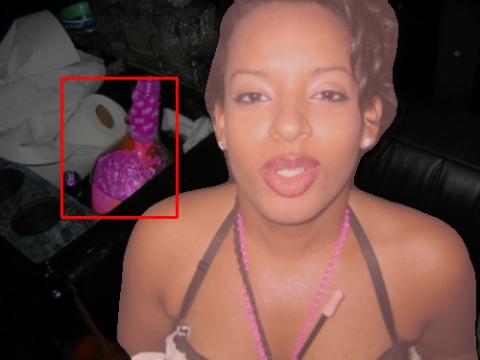}    	
\end{subfigure}
\begin{subfigure}{0.2\textwidth}
	\includegraphics[width=1\textwidth]{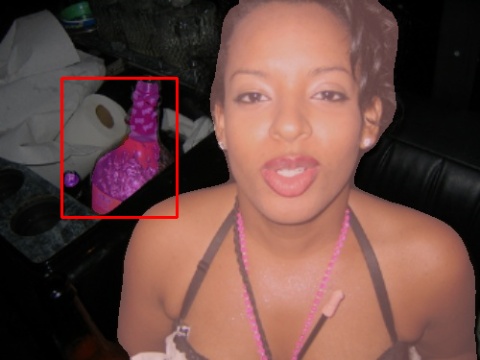} 
\end{subfigure}
\par
\begin{subfigure}{0.2\textwidth}\centering	\small
	\includegraphics[width=1\textwidth]{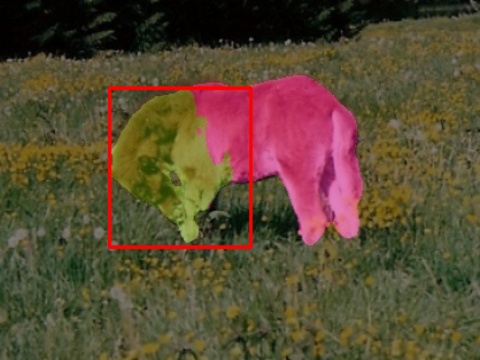}
	(a) T-FPN 
\end{subfigure}
\begin{subfigure}{0.2\textwidth}\centering	\small
	\includegraphics[width=1\textwidth]{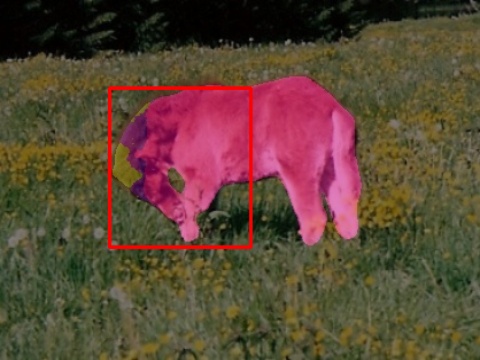}
	(b) MSwin-P 
\end{subfigure}
\begin{subfigure}{0.2\textwidth}\centering \small	
	\includegraphics[width=1\textwidth]{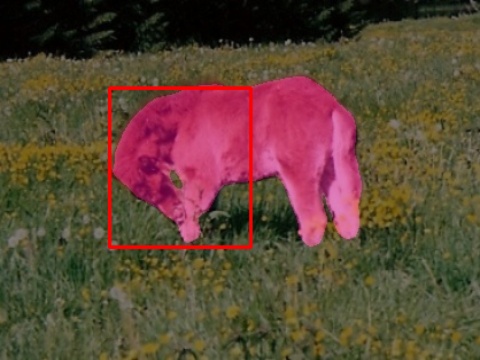}
	(c) MSwin-S
\end{subfigure}
\begin{subfigure}{0.2\textwidth}\centering	\small
	\includegraphics[width=1\textwidth]{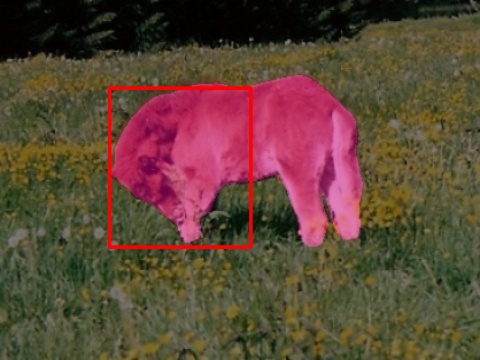}
	(d) MSwin-C
\end{subfigure}	
\caption{Segmentation examples on the VOC2012 validation set.}
\label{FIG:VOC}
\end{figure*}

We then experimented the segmentation models with Swin-S and Swin-B as backbones to verify the effectiveness of the baseline T-FPN, as well as three MSwin models on this dataset. The mIoU scores on the validation set are summarized in Table \ref{tab:voc_val}, where SS and MS are the abbreviates of single-scale and multi-scale prediction for mIoU scores, respectively. From the table, we can see that even with a softmax classifier as a decoder, the T-FPN backed on Swin Transformer is a powerful semantic segmentation model. Applying the small-sized backbone Swin-S, the proposed MSwin-P, MSwin-S and MSwin-C further improve the baseline T-FPN by 0.89\%, 1.28\% and 0.71\%, and by 0.88\%, 0.67\% and 0.63\% in terms of mIoU score when using single-scale and multi-scale predictions, respectively. When applying the more powerful backbone Swin-B, the three MSwin models further improve the performance of the baseline similarly. Figure \ref{FIG:VOC} gives some visualization results of the four segmentation methods on the validation dataset.  

We mixed the {\em train+val} datasets, fine-tuned the models, and used the multi-scale prediction on the test set, then uploaded the results to the evaluation server. In Table \ref{tab:voc_test}, we compared our models with the recently proposed methods. The three MSwin models achieve the new state-of-the-arts without the pre-training on MS COCO dataset.

\begin{table}[t]
  \centering
  \begin{tabular}{|c|c|c|c|}
    \hline
    Method & Backbone & SS & MS \\
    \hline
    T-FPN & Swin-S &80.69 &82.07  \\
    MSwin-P & Swin-S &81.58 &82.95 \\
    MSwin-S & Swin-S &81.97 &82.74 \\
    MSwin-C & Swin-S &81.40 &82.70 \\
    T-FPN & Swin-B &81.81 &83.58\\
    MSwin-P & Swin-B &82.85 &83.82 \\
    MSwin-S & Swin-B &82.86 &84.27 \\
    MSwin-C & Swin-B &83.50 &84.47 \\
    \hline
  \end{tabular}
  \caption{Results on PASCAL VOC2012 validation set.}
  \label{tab:voc_val}
\end{table}

\begin{table*}[t]
\centering
  \begin{tabular}{|c|c|c|c|}
    \hline
    Method & Backbone &COCO & Score  \\
    \hline
    DANet\cite{CVPR19:DANET} &ResNet-101 &\xmark  & 82.6 \\
    DeepLab v3+\cite{ECCV18:DEEPLAB} &Xception-71 &\cmark  & 87.8 \\
    Auto-DeepLab-L\cite{CVPR19:AUTODEEPLAB} &- &\cmark &85.6 \\
    EncNet\cite{CVPR18:ENCNET} &ResNet-101 &\xmark  & 82.9 \\
	APCNet\cite{CVPR19:APCNET} &ResNet-101 &\xmark  & 84.2 \\
	PSANet\cite{ECCV18:PSANET} &ResNet-101 &\cmark &85.7 \\
	EMANet\cite{ICCV19:EMANET} &ResNet-101 &\cmark &87.7 \\  
    \hline
    MSwin-P & Swin-B & \xmark & 85.9  \\
    MSwin-S & Swin-B & \xmark & 86.1  \\
    MSwin-C & Swin-B & \xmark & 85.7 \\
    \hline
  \end{tabular}
  \caption{Online evaluation of PASCAL VOC 2012 test set.}
  \label{tab:voc_test}
\end{table*}

\subsection{Results on COCO-Stuff 10K dataset}
We conducted experiments on the COCO-Stuff 10K dataset to prove the generalization ability of the proposed methods. All models were optimized by 40,000 iterations. The comparisons with previously state-of-the-arts are reported in Table \ref{tab:cocostuff_val}. Our pure Transformer-based models, including the baseline T-FPN, consistently outperform all the ConvNet-based counterparts. When applying three different aggregation strategies in the decoder and multi-scale prediction, MSwin-P, MSwin-S and MSwin-C further boost the baseline T-FPN by 0.3\%, 1.0\% and 0.5\% in terms of mIoU score, respectively. Our models achieve a slightly lower mIoU compared to the OCRNet backend on HRFormer-B on this dataset.

\begin{table*}[t]
\centering
  \begin{tabular}{|c|c|c|c|}
    \hline
    Method & Backbone & SS & MS \\
    \hline
    DANet\cite{CVPR19:DANET} &ResNet-101 &- &39.7 \\
    OCRNet\cite{ECCV20:OCR} &HRNetV2-48 &- &40.5\\
    ACNet\cite{ICCV19:ACNET} &ResNet-101 &- &40.1 \\
    EMANet\cite{ICCV19:EMANET} &ResNet-101 &- &39.9 \\
    RegionContrast\cite{ICCV21:RC} &ResNet-101 &- &40.7 \\
    OCRNet\cite{NIPS21:HRFORMER} & HRFormer-B &- &43.3 \\
    \hline
    T-FPN & Swin-S &39.6 &41.1 \\
    MSwin-P & Swin-S &39.8 &41.4 \\
    MSwin-S & Swin-S &40.2 &42.1 \\
    MSwin-C & Swin-S &39.6 &41.6 \\
    T-FPN & Swin-B &40.9 &41.7 \\
    MSwin-P & Swin-B &41.3 &42.7 \\
    MSwin-S & Swin-B &41.1 &42.4 \\
    MSwin-C & Swin-B &41.0 &42.8 \\
    \hline
  \end{tabular}
  \caption{mIoU results on the COCO-Stuff 10K test set.}
  \label{tab:cocostuff_val}
\end{table*}


\subsection{Results on ADE20K dataset}

\begin{figure*}[t]\centering
\begin{subfigure}{0.2\textwidth}
	\includegraphics[width=1\textwidth]{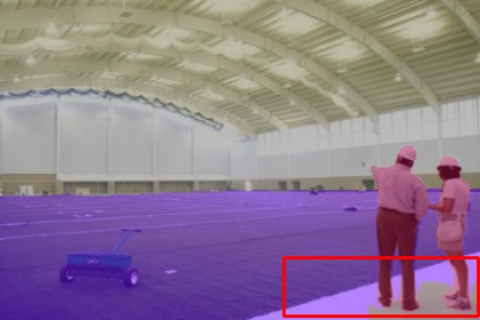}
\end{subfigure}
\begin{subfigure}{0.2\textwidth}
	\includegraphics[width=1\textwidth]{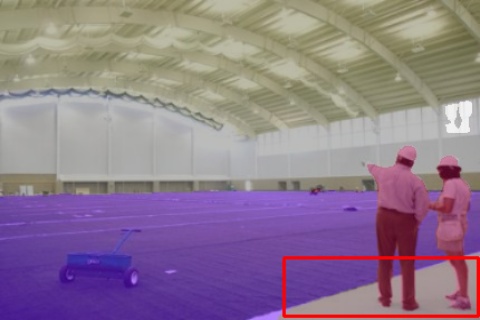}
\end{subfigure}
\begin{subfigure}{0.2\textwidth}
	\includegraphics[width=1\textwidth]{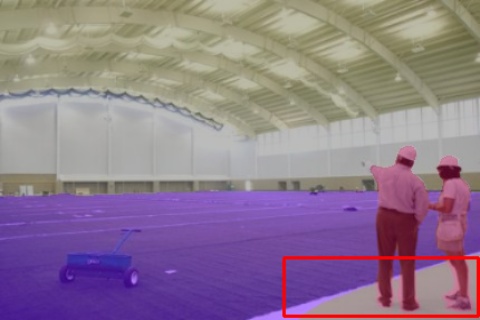}
\end{subfigure}
\begin{subfigure}{0.2\textwidth}\centering
	\includegraphics[width=1\textwidth]{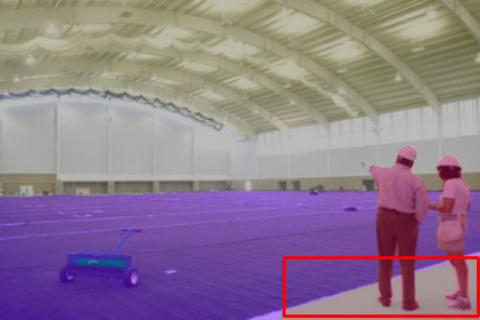}
\end{subfigure}	
\par
\begin{subfigure}{0.2\textwidth}
	\includegraphics[width=1\textwidth]{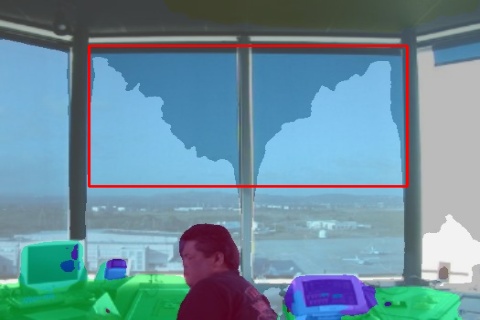}
\end{subfigure}
\begin{subfigure}{0.2\textwidth}
	\includegraphics[width=1\textwidth]{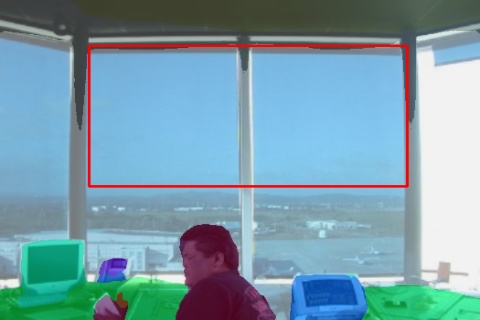}
\end{subfigure}
\begin{subfigure}{0.2\textwidth}
	\includegraphics[width=1\textwidth]{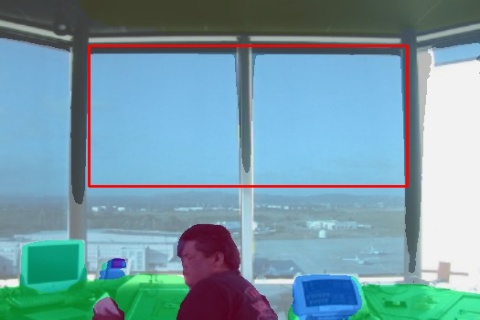}
\end{subfigure}
\begin{subfigure}{0.2\textwidth}\centering
	\includegraphics[width=1\textwidth]{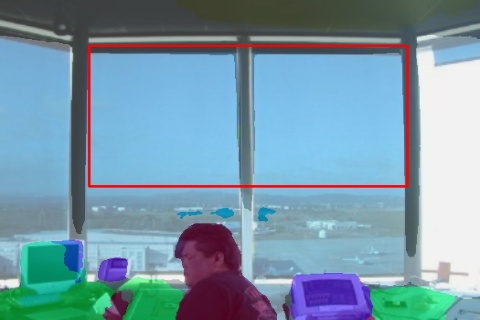}
\end{subfigure}	
\par
\begin{subfigure}{0.2\textwidth}
	\includegraphics[width=1\textwidth]{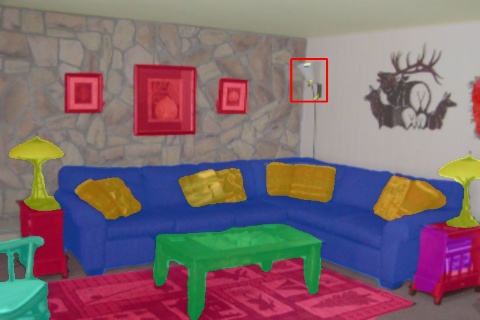}
\end{subfigure}	
\begin{subfigure}{0.2\textwidth}
	\includegraphics[width=1\textwidth]{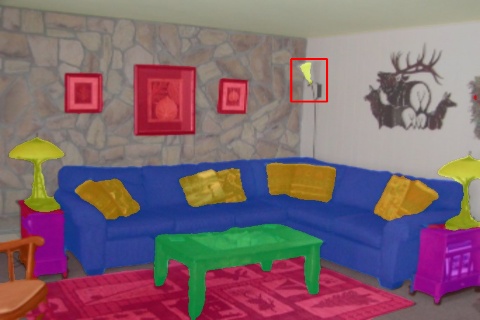}
\end{subfigure}
\begin{subfigure}{0.2\textwidth}
	\includegraphics[width=1\textwidth]{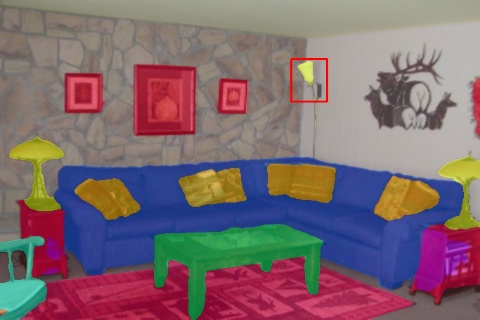}
\end{subfigure}
\begin{subfigure}{0.2\textwidth}\centering	
	\includegraphics[width=1\textwidth]{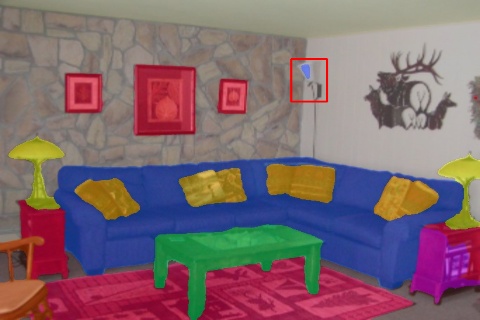}
\end{subfigure}
\par
\begin{subfigure}{0.2\textwidth}	
	\includegraphics[width=1\textwidth]{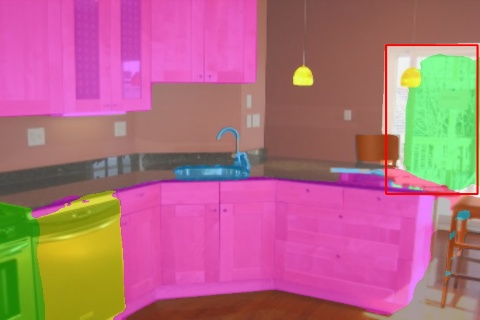}    	
\end{subfigure}
\begin{subfigure}{0.2\textwidth}	
	\includegraphics[width=1\textwidth]{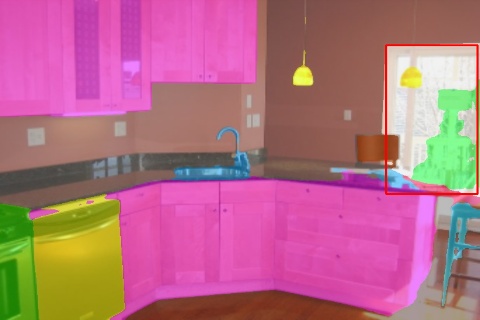}    
\end{subfigure}
\begin{subfigure}{0.2\textwidth}	
	\includegraphics[width=1\textwidth]{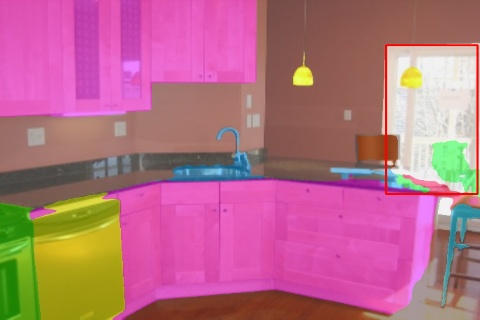}    
\end{subfigure}
\begin{subfigure}{0.2\textwidth}
	\includegraphics[width=1\textwidth]{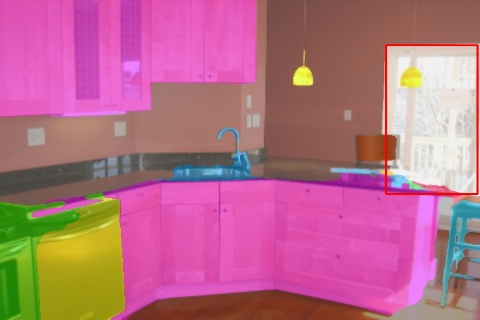} 
\end{subfigure}
\par
\begin{subfigure}{0.2\textwidth}\centering	
	\includegraphics[width=1\textwidth]{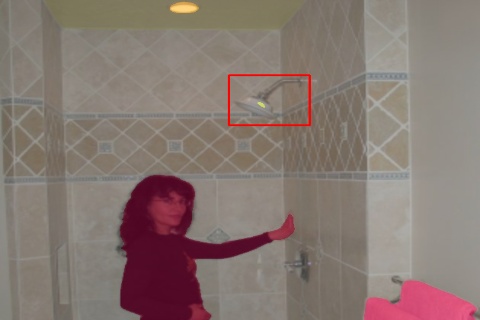}
	(a) T-FPN 
\end{subfigure}
\begin{subfigure}{0.2\textwidth}\centering	
	\includegraphics[width=1\textwidth]{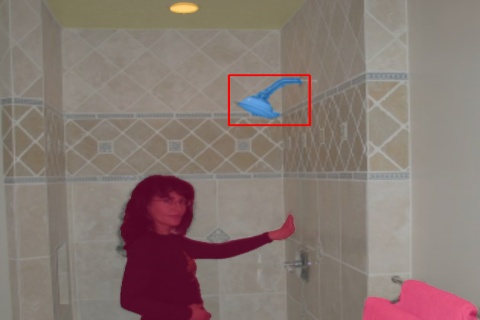}
	(b) MSwin-P 
\end{subfigure}
\begin{subfigure}{0.2\textwidth}\centering 
	\includegraphics[width=1\textwidth]{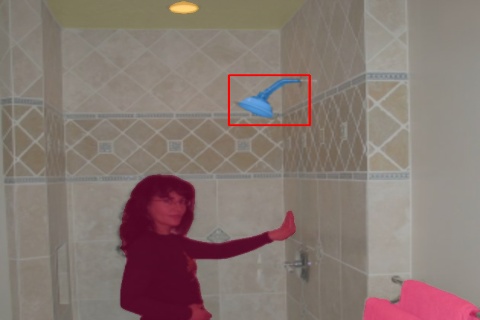}
	(c) MSwin-S
\end{subfigure}
\begin{subfigure}{0.2\textwidth}\centering	
	\includegraphics[width=1\textwidth]{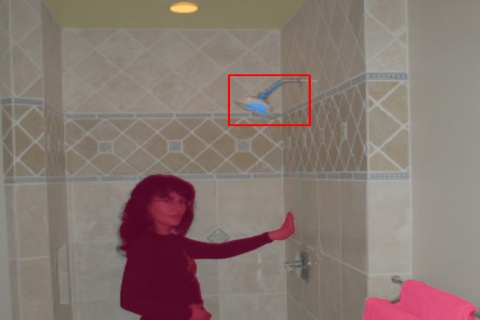}
	(d) MSwin-C
\end{subfigure}	
\caption{Segmentation examples on the ADE20K validation set.}
\label{FIG:ADE20K}
\end{figure*}

On this dataset, we used both Swin-S and Swin-B as backbones for model training to evaluate the performance. Table \ref{tab:ade_val} shows the comparisons of FLOPs, as well as the results on the validation set. We can observe that the Transformer-based segmentation models generally needs more computational resources, except the recently proposed SegFormer. However, backend with Swin Transformer, our models are more computationally efficient. Applying the self-attention on multi-shifted windows in the decoder doubles the FLOPs in MSwin compared to T-FPN, but leads to more accurate segmentation results. On the validation dataset, all Transformer-based segmentation models substantially outperform the ConvNet counterparts by a large margin. The reason for this is mainly because the self-attention mechanism in Transformers has a very strong capability in modelling the spatial dependency, which can effectively parse very complex scenes in different visual environments by capturing the long-range dependencies. On the validation set, our MSwin models achieve comparable results to UPerNet backend on Swin Transformer, SETR backend on ViT-L and SegFormer. Some segmentation examples are illustrated in Figure \ref{FIG:ADE20K}.

We then fine-tuned the MSwin-P model on the {\em train+validation} set to improve the overall classification accuracy. On the evaluation server, MSwin-P obtains the pixel-wise accuracy of 0.767, mIoU 0.457 and the final test score of 0.612. 

\begin{table*}[t]
\centering
  \begin{tabular}{|c|c|c|c|c|}
    \hline
    Method & Backbone & FLOPs &SS & MS \\
	\hline
	EncNet \cite{CVPR18:ENCNET} &ResNet-101 & 219G &- &44.65 \\   	
	ACNet \cite{ICCV19:ACNET} &ResNet-101 &- &- &45.90 \\
	APCNet \cite{CVPR19:APCNET} &ResNet-101 &282G &- &45.38 \\
	OCRNet \cite{ECCV20:OCR} &HRNetV2-48 &165G &- &45.66 \\
	UPerNet \cite{ECCV18:UPERNET} &ResNet-50 &- &41.22 &- \\
	RegionContrast\cite{ICCV21:RC} &ResNet-101 &- & - & 46.9 \\
	UPerNet \cite{ICCV21:SWIN} & Swin-B &300G & 48.35 & 49.65 \\
	UPerNet \cite{NIPS21:FOCAL} & Focal-B &342G & 49.00 & 50.50 \\
	FPN \cite{ICCV21:PVT} &PVT-Large & 80G & 42.10 &44.80 \\	
	SETR \cite{CVPR21:SETR} &ViT-L &270G  &48.64 &50.28 \\ 
	SegFormer \cite{NIPS21:SEGFORMER} &MiT-B5 &183G &49.13 &50.22 \\
	DPT \cite{ICCV21:DPT} & ViT-Hybrid &- &- &49.02 \\
	OCRNet\cite{NIPS21:HRFORMER} & HRFormer-B &283G &- &50.00 \\
    \hline
    T-FPN & Swin-S &87G  & 46.38 &48.15  \\
    MSwin-P & Swin-S &230G &47.11 &48.55 \\
    MSwin-S & Swin-S &247G &47.52 &48.56 \\
    MSwin-C & Swin-S &220G &46.26 &48.12 \\
    T-FPN & Swin-B &127G &47.70 & 49.41 \\
    MSwin-P & Swin-B &270G &48.38 &50.29 \\
    MSwin-S & Swin-B &288G &48.54 &50.26 \\
    MSwin-C & Swin-B &260G &48.70 &50.13 \\
    \hline
  \end{tabular}
  \caption{Results on ADE20K validation set.}
  \label{tab:ade_val}
\end{table*}

\subsection{Results on Cityscapes dataset}

\begin{figure*}[t]\centering
\begin{subfigure}{0.24\textwidth}
	\includegraphics[width=1\textwidth]{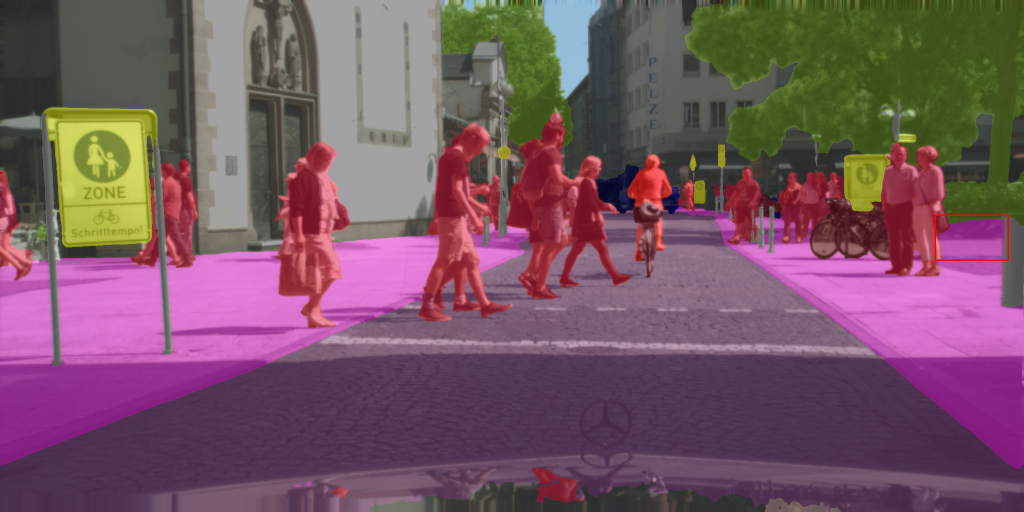}
\end{subfigure}
\begin{subfigure}{0.24\textwidth}
	\includegraphics[width=1\textwidth]{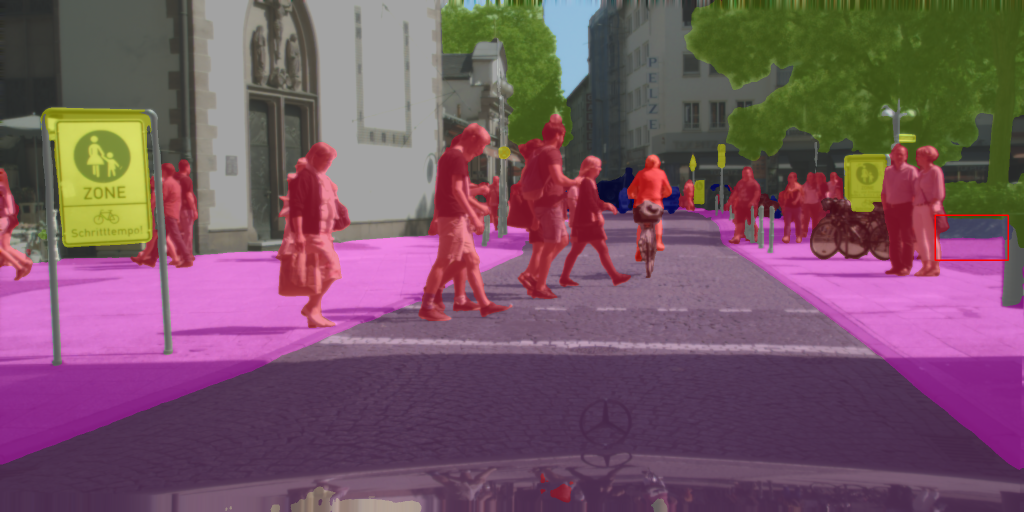}
\end{subfigure}
\begin{subfigure}{0.24\textwidth}
	\includegraphics[width=1\textwidth]{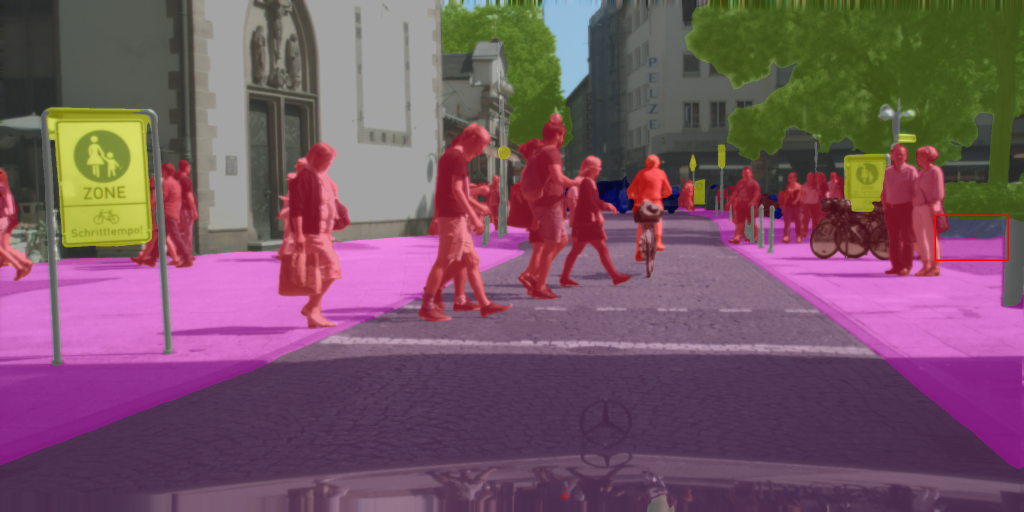}
\end{subfigure}
\begin{subfigure}{0.24\textwidth}\centering
	\includegraphics[width=1\textwidth]{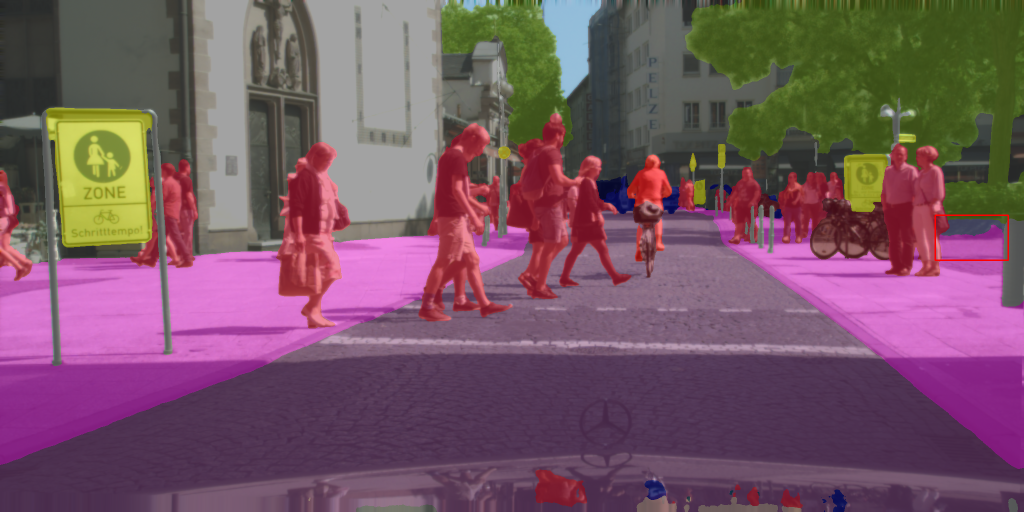}
\end{subfigure}	
\par
\begin{subfigure}{0.24\textwidth}
	\includegraphics[width=1\textwidth]{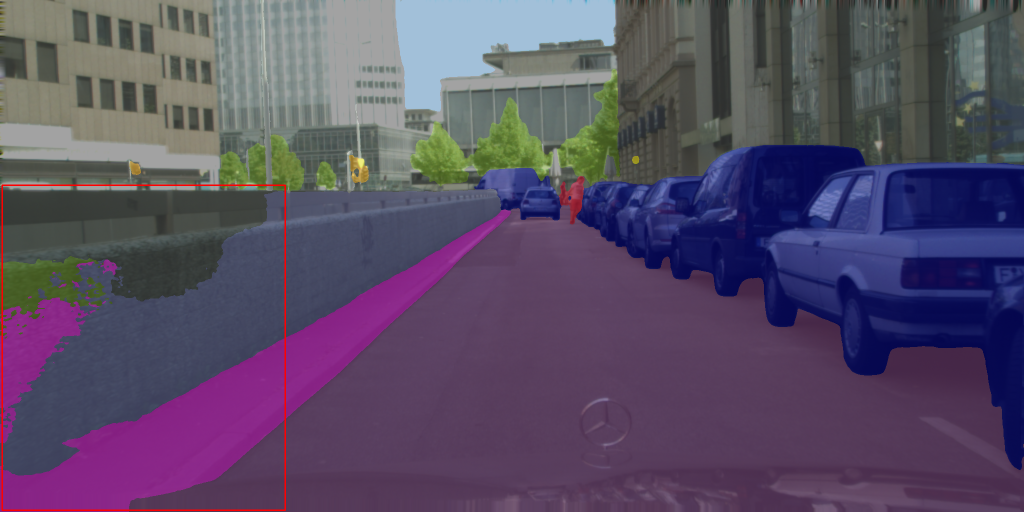}
\end{subfigure}	
\begin{subfigure}{0.24\textwidth}
	\includegraphics[width=1\textwidth]{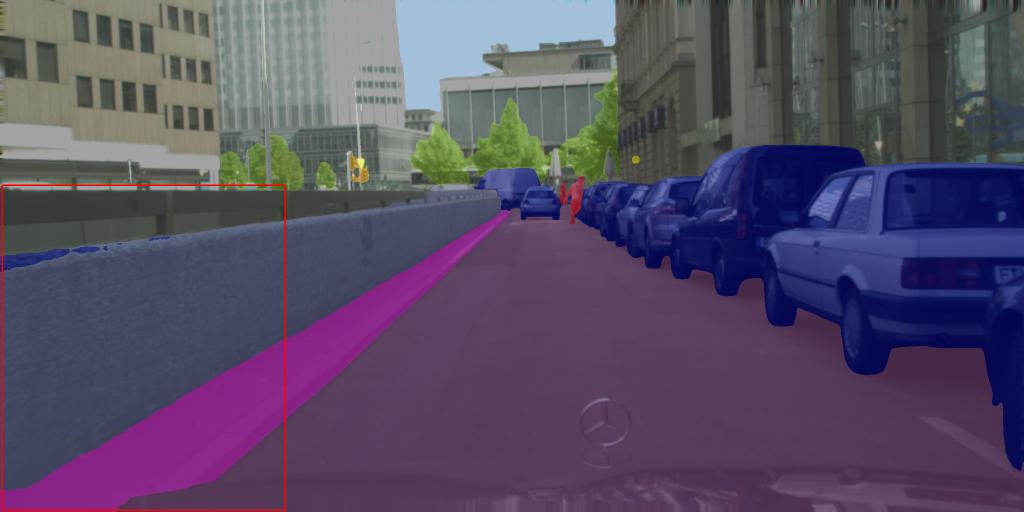}
\end{subfigure}
\begin{subfigure}{0.24\textwidth}
	\includegraphics[width=1\textwidth]{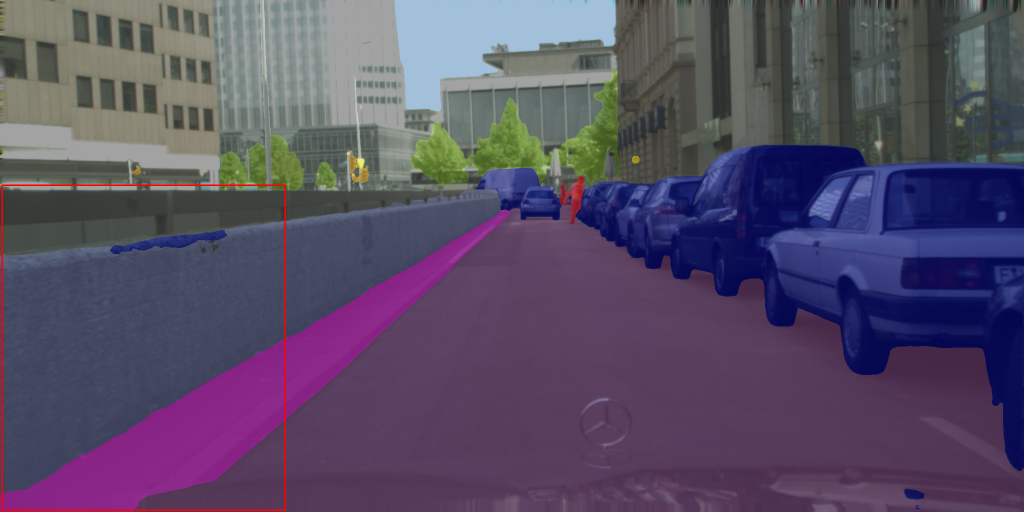}
\end{subfigure}
\begin{subfigure}{0.24\textwidth}\centering	
	\includegraphics[width=1\textwidth]{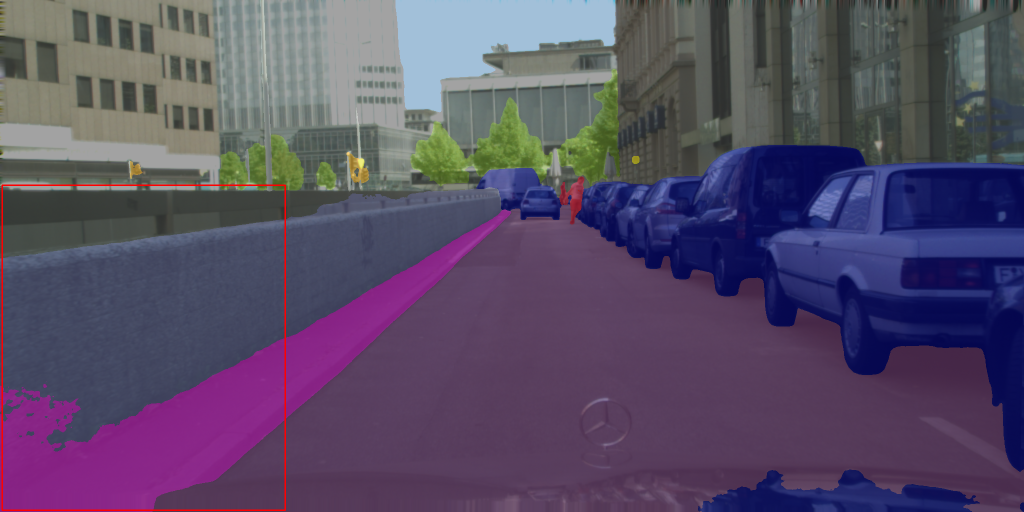}
\end{subfigure}
\par
\begin{subfigure}{0.24\textwidth}	
	\includegraphics[width=1\textwidth]{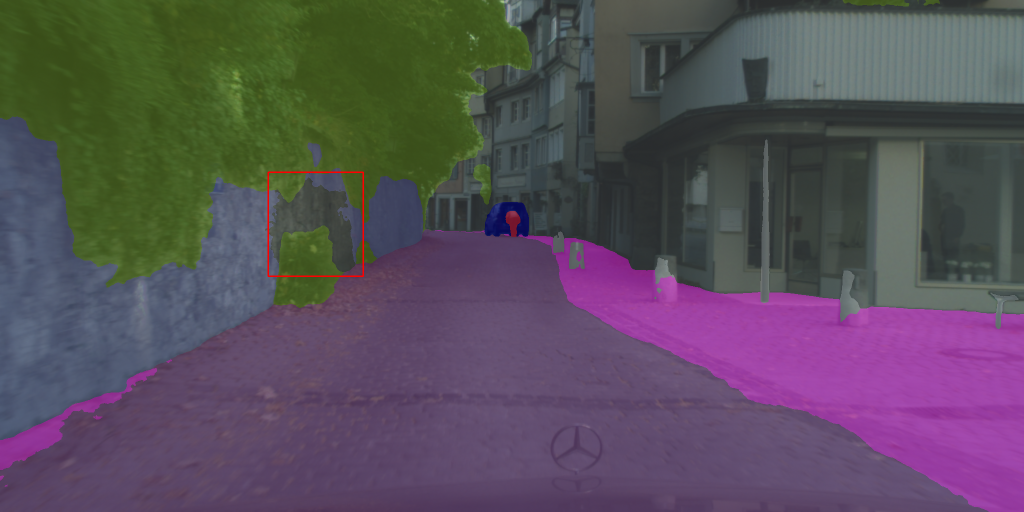}    	
\end{subfigure}
\begin{subfigure}{0.24\textwidth}	
	\includegraphics[width=1\textwidth]{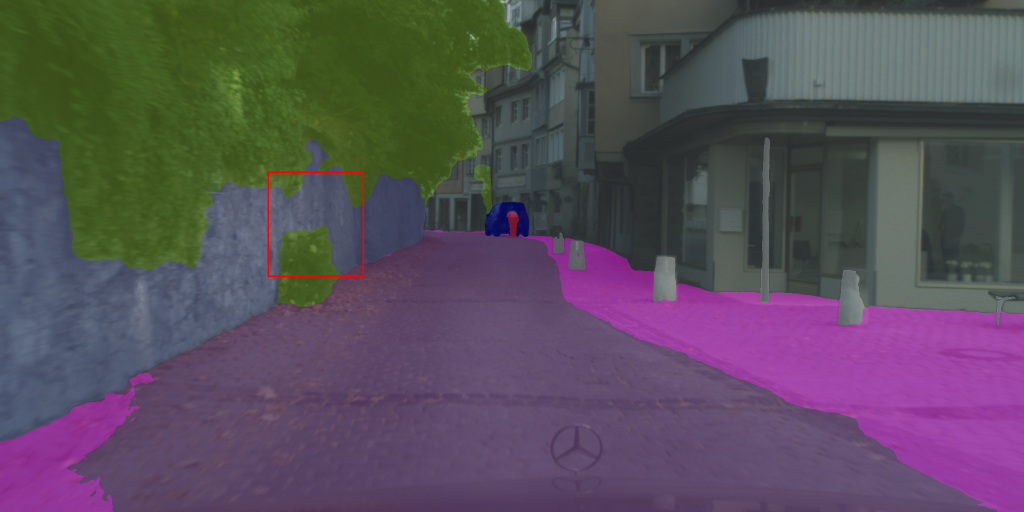}    	
\end{subfigure}
\begin{subfigure}{0.24\textwidth}	
	\includegraphics[width=1\textwidth]{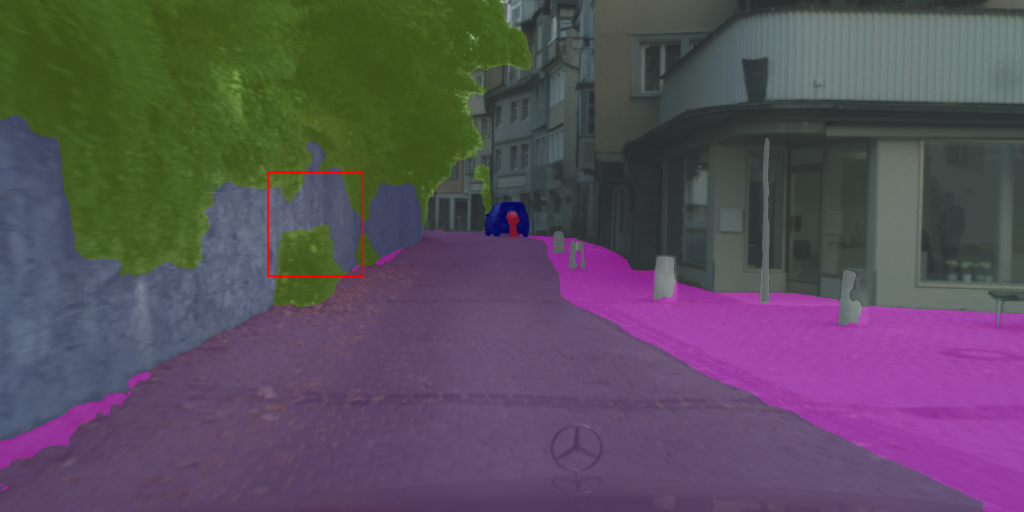}    	
\end{subfigure}
\begin{subfigure}{0.24\textwidth}\centering	
	\includegraphics[width=1\textwidth]{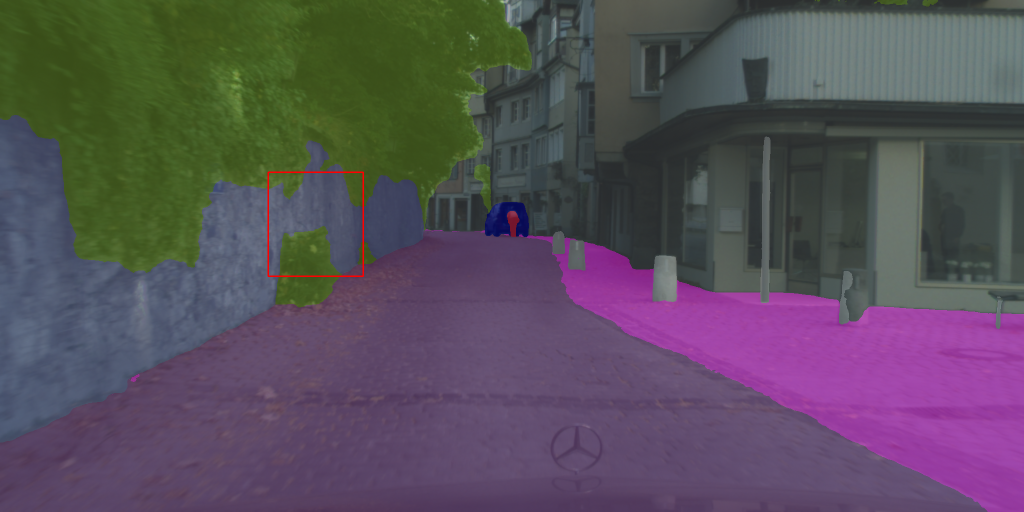} 
\end{subfigure}
\par
\begin{subfigure}{0.24\textwidth}\centering \small
	\includegraphics[width=1\textwidth]{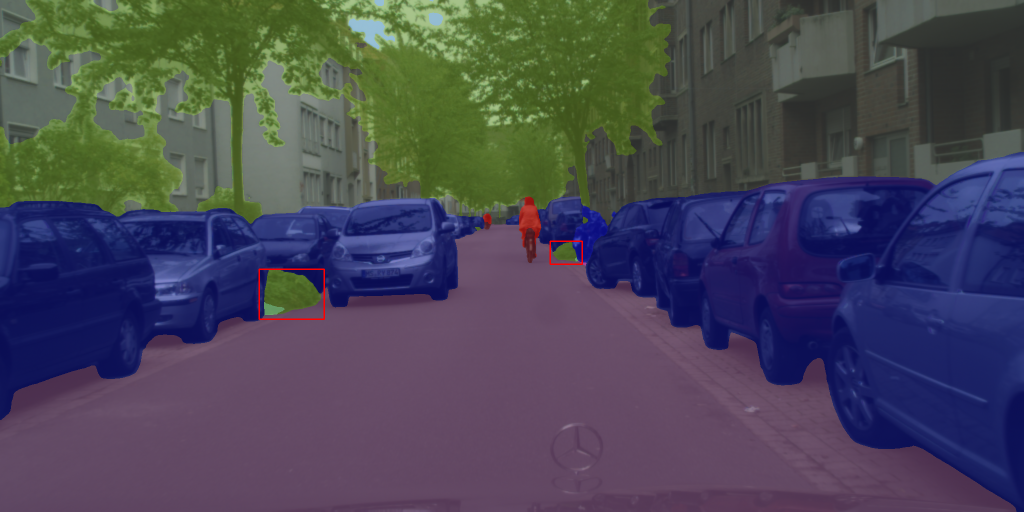}
	(a) T-FPN 
\end{subfigure}
\begin{subfigure}{0.24\textwidth}\centering \small
	\includegraphics[width=1\textwidth]{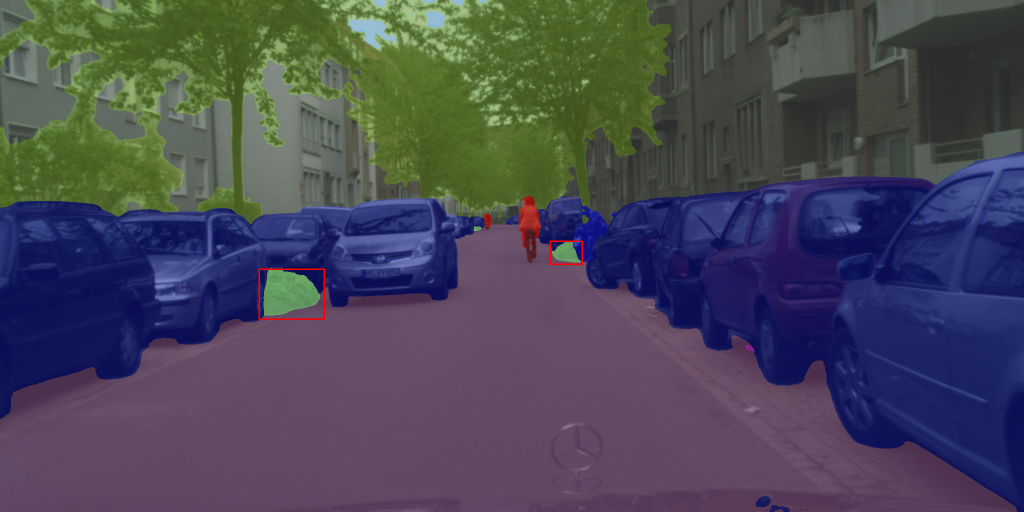}
	(b) MSwin-P 
\end{subfigure}
\begin{subfigure}{0.24\textwidth}\centering \small
	\includegraphics[width=1\textwidth]{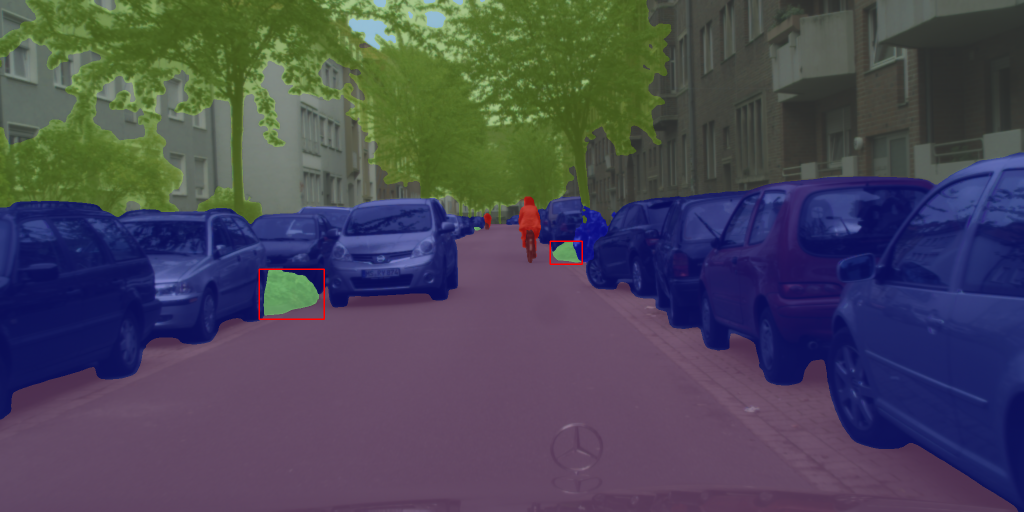}
	(c) MSwin-S
\end{subfigure}
\begin{subfigure}{0.24\textwidth}\centering \small	
	\includegraphics[width=1\textwidth]{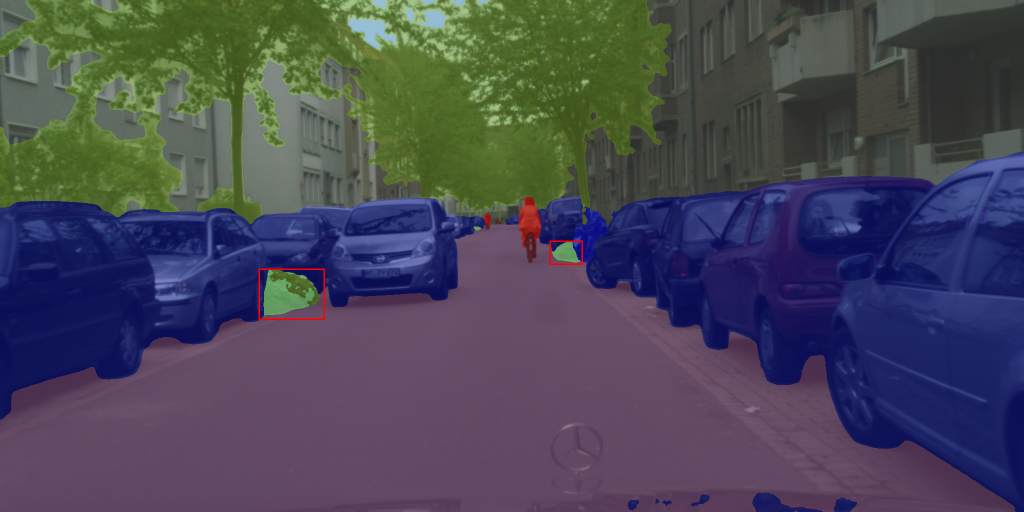}
	(d) MSwin-C
\end{subfigure}	
\caption{Scene segmentation examples on the Cityscapes validation set.}
\label{FIG:CITYSCAPES}
\end{figure*}

\begin{figure*}[t]\centering
\begin{subfigure}{0.24\textwidth}\centering \small
	\includegraphics[width=1\textwidth]{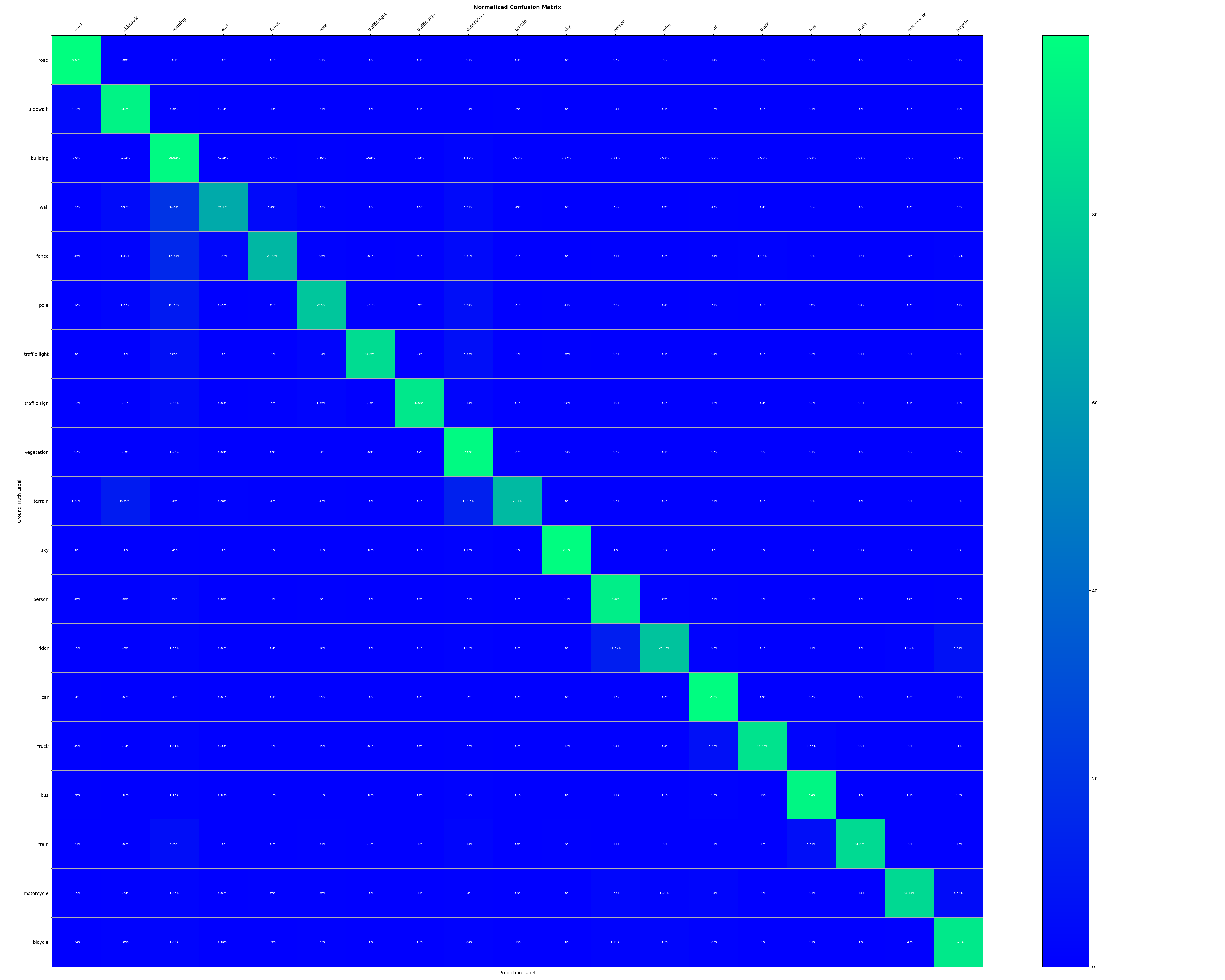}
	(a) T-FPN 
\end{subfigure}
\begin{subfigure}{0.24\textwidth}\centering \small
	\includegraphics[width=1\textwidth]{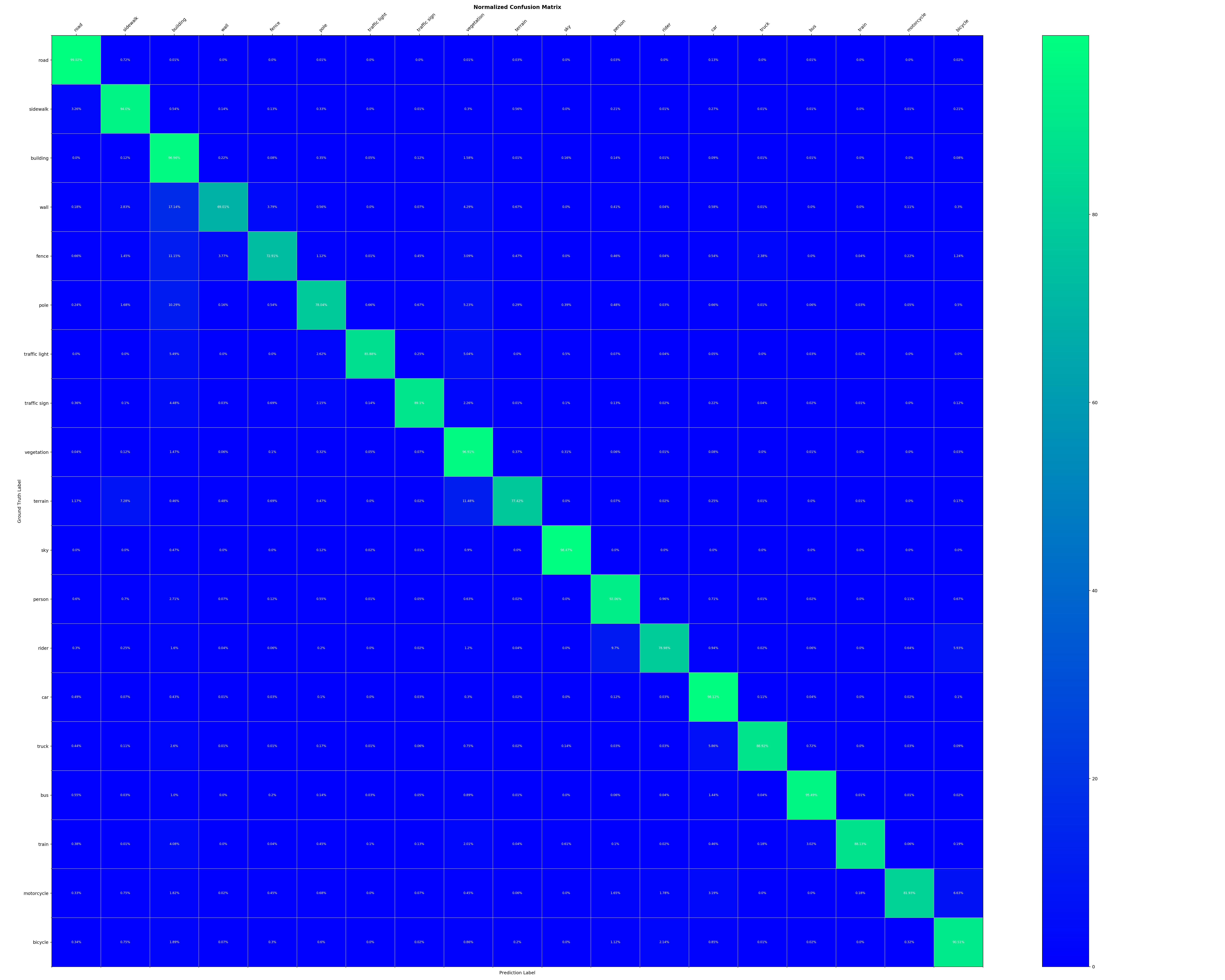}
	(b) MSwin-P 
\end{subfigure}
\begin{subfigure}{0.24\textwidth}\centering \small
	\includegraphics[width=1\textwidth]{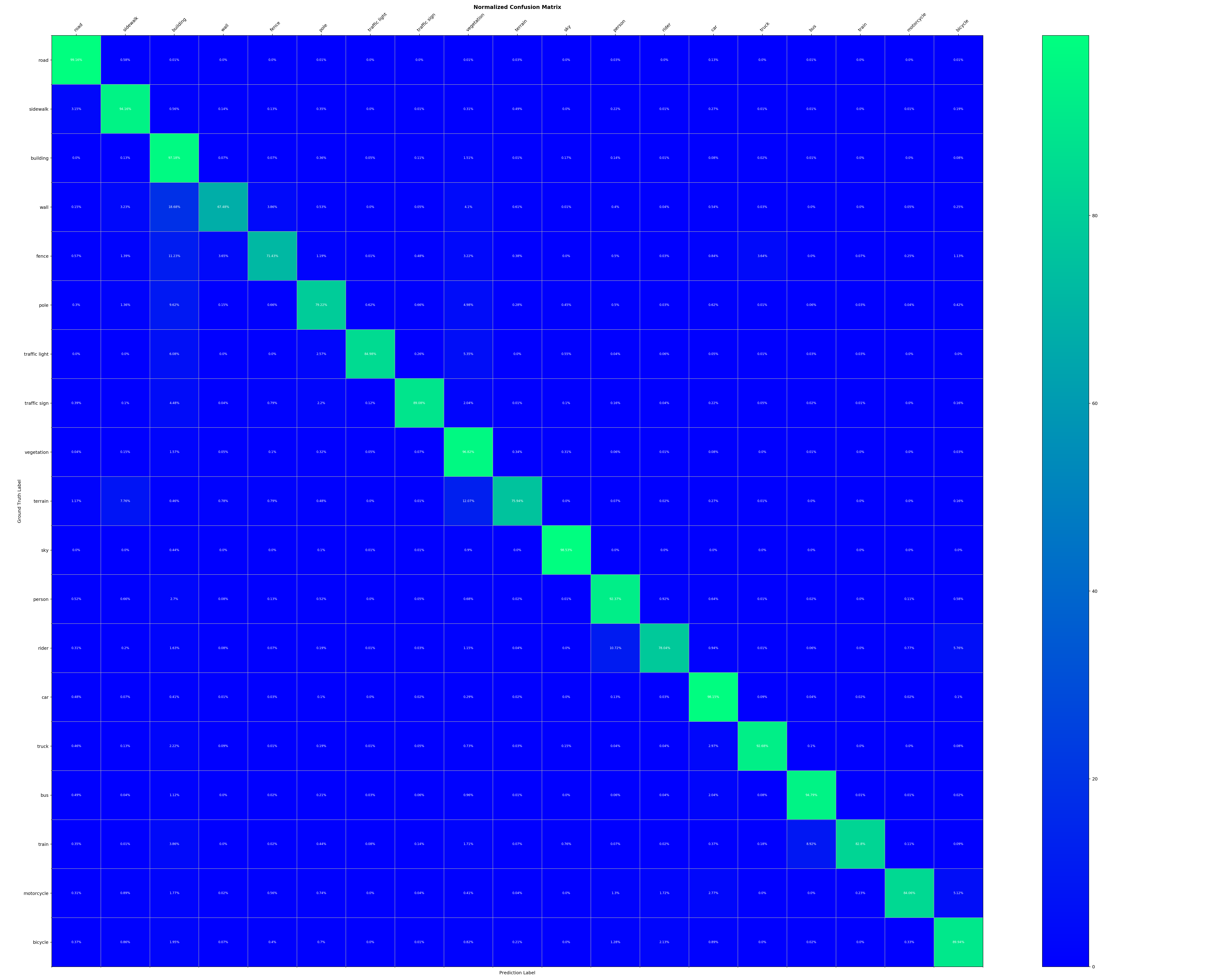}
	(c) MSwin-S
\end{subfigure}
\begin{subfigure}{0.24\textwidth}\centering \small	
	\includegraphics[width=1\textwidth]{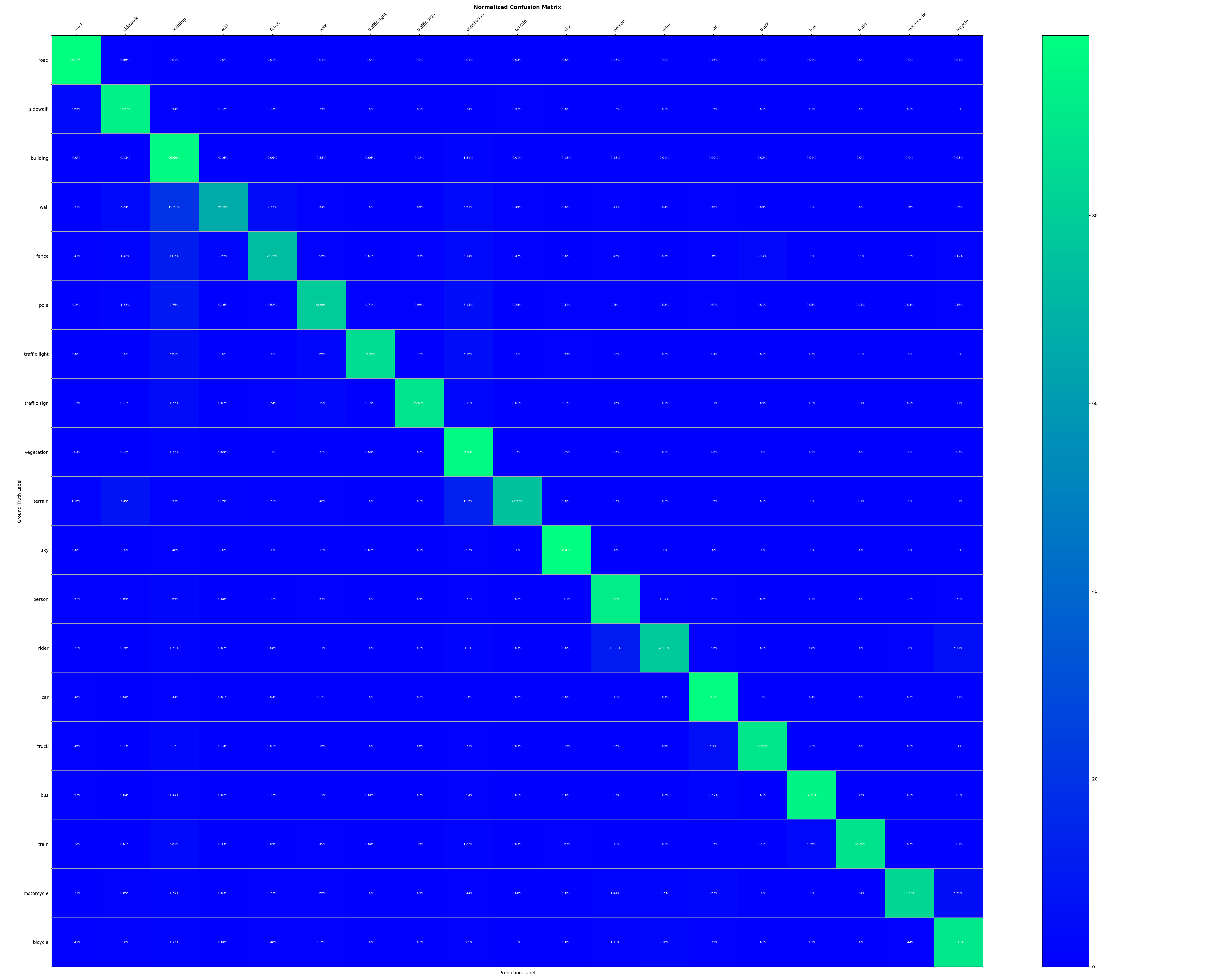}
	(d) MSwin-C
\end{subfigure}	
\caption{Confusion matrices on the Cityscapes validation set.}
\label{FIG:CFM}
\end{figure*}

On this dataset, we directly used Swin-B as the backbone to evaluate the effectiveness of the proposed MSwin models in street scene segmentation. Only using the fine-labeled dataset, we show the improvements brought by the decoder with self-attention on multi-shifted windows based on the T-FPN encoder in Table \ref{tab:cityscapes_val}. On the validation set, the parallel structure of MSwin achieves the best mIoU score when using single-scale prediction, which improves mIoU by 0.67\% compared to T-FPN. Applying the sequential structure, the model obtains the best performance with multi-scale prediction, which outperforms the baseline by 0.62\% of mIoU. We show some segmentation examples and the confusion matrices of dense prediction in Figure \ref{FIG:CITYSCAPES} and \ref{FIG:CFM}, respectively. Compared with a {\em thin} decoder based on T-FPN, self-attention on multi-shifted windows can further reduce false positives and lead to finer details of street scenes.

\begin{table}[t]
\centering
  \begin{tabular}{|c|c|c|c|}
    \hline
    Method & Backbone & SS & MS \\
    \hline
    T-FPN & Swin-B &80.39 &81.77 \\
    MSwin-P & Swin-B &81.06 &82.10 \\
    MSwin-S & Swin-B &80.87 &82.39 \\
    MSwin-C & Swin-B &80.78 &82.04 \\
    \hline
  \end{tabular}
  \caption{Results on Cityscapes validation set.}
  \label{tab:cityscapes_val}
\end{table}

We pre-trained MSwin-S and MSwin-C on the coarsely labeled training data then fine-tuned them on the fine labeled training dataset. After that, we applied the multi-scale prediction on the test set and submitted the results to the evaluation server. The overall comparisons of our models with some recently proposed methods are summarized in Table \ref{tab:cityscapes_test}. We can see that without the pre-training with the coarsely labeled images, the proposed MSwin-P obtains the mIoU score 81.8\%, which is slightly worse than the best scene segmentation models based on ConvNets. When pre-trained with the coarsely labeled data, MSwin-S and MSwin-C obtain the best results. Specifically, compared to the Transformer-based scene segmentation models SETR and SegFormer, MSwin-S and MSwin-C outperform the second-best by 0.5\% and 0.3\% in terms of mIoU score, respectively.

\begin{table*}[t]
\centering
  \begin{tabular}{|c|c|c|c|}
    \hline
    Method & Backbone  & Coarse & mIoU  \\
    \hline
    DenseASPP \cite{CVPR18:DENSEASPP} &DenseNet-161 &\xmark &80.6 \\
    DeepLab V3+ \cite{ECCV18:DEEPLAB}& Xception-71 &\cmark &82.1 \\
    Auto-DeepLab-L \cite{CVPR19:AUTODEEPLAB}& - &\xmark &80.4 \\
    DANet \cite{CVPR19:DANET}	& ResNet-101 &\xmark &81.5 \\
    HANet \cite{CVPR20:HANET} & ResNet-101 &\xmark &80.9  \\
    HRNetV2 \cite{TPAMI:HRNET}& HRNetV2-48 &\xmark &81.6  \\
    OCRNet \cite{ECCV20:OCR} &HRNetV2-48 &\xmark  &82.4 \\
    RegionContrast\cite{ICCV21:RC} &ResNet-101 &\xmark  &82.3 \\
    ACNet\cite{ICCV19:ACNET} & ResNet-101 &\xmark &82.3  \\
    SETR \cite{CVPR21:SETR} & ViT-L  &\xmark  &81.1 \\
    SETR \cite{CVPR21:SETR} & ViT-L  &\cmark  &81.6 \\
    SegFormer \cite{NIPS21:SEGFORMER} & MiT-B5  &\cmark & 82.2 \\
    OCRNet \cite{NIPS21:HRFORMER} & HRFormer  &\cmark & 82.6 \\
    \hline
    MSwin-P & Swin-B  & \xmark & 81.8 \\
	MSwin-S & Swin-B  & \cmark & 82.9 \\	
	MSwin-C & Swin-B  & \cmark & 82.7 \\    
    \hline
  \end{tabular}
  \caption{Online evaluation of Cityscapes test set.}
  \label{tab:cityscapes_test}
\end{table*}

\section{Conclusion}
\label{SEC:CONCLUSION}
In this work, we have presented a convolution-free deep scene parsing model based on Swin Transformers. Different from the existing FCN based semantic segmentation models that parse complex semantic contextual information by enlarging receptive fields with dilated convolutions, the Transformer-based models consider the spatial and semantic dependencies. We have analyzed the computational properties of Swin Transformer, and designed a feature pyramid encoder that aggregates multiple-layer outputs without modifying the backbone. Furthermore, we have proposed the self-attention on multiple shifted windows to diversify the feature representations in the decoder. Extensive experiments on four public benchmarks demonstrate our models set new state-of-the-art performance.

\newpage
\bibliographystyle{named}
\bibliography{ref}

\end{document}